\pdfoutput=1
\documentclass[11pt]{article}

\usepackage{EMNLP2023}

\usepackage{times}
\usepackage{latexsym}

\usepackage[T1]{fontenc}
\usepackage[utf8]{inputenc}

\usepackage{microtype}

\usepackage{inconsolata}

\usepackage{microtype}
\usepackage{graphicx}
\usepackage{booktabs} % for professional tables

\usepackage{hyperref}

\usepackage{amsmath}
\usepackage{amssymb}
\usepackage{mathtools}
\usepackage{amsthm}

\usepackage[utf8]{inputenc} % allow utf-8 input
\usepackage[T1]{fontenc}    % use 8-bit T1 fonts
\usepackage{url}            % simple URL typesetting
\usepackage{booktabs}       % professional-quality tables
\usepackage{amsfonts}       % blackboard math symbols
\usepackage{nicefrac}       % compact symbols for 1/2, etc.
\usepackage{microtype}      % microtypography
\usepackage{xcolor}         % colors
\usepackage{wrapfig}
\usepackage[normalem]{ulem}
\definecolor{darker}{rgb}{0,0.15,0.7}
\usepackage{amsmath}
\usepackage{amssymb}
\usepackage{mathtools}
\usepackage{amsthm}
\usepackage[capitalize,noabbrev]{cleveref}

\usepackage{xurl}
\usepackage{latexsym}
\usepackage{booktabs}
\usepackage{multirow}
\usepackage[T1]{fontenc}
\usepackage{graphicx}
\usepackage{graphicx}
\usepackage{soul}
\usepackage{multirow}
\usepackage{bbold}
\usepackage{makecell}
\usepackage{amsmath}
\usepackage{hyperref}       % hyperlinks
\usepackage{cleveref}
\usepackage{bm}
\usepackage{xspace}
\usepackage{adjustbox}
\usepackage{caption}
\usepackage{subcaption}
\usepackage[utf8]{inputenc}
\usepackage{enumitem}

\usepackage[capitalize,noabbrev]{cleveref}
\usepackage{microtype}

\usepackage{tcolorbox}

\theoremstyle{plain}

\theoremstyle{definition}

\theoremstyle{remark}

\usepackage[textsize=tiny]{todonotes}

\newcommand{\method}{\textsc{Retro}\xspace}
\newcommand{\methodplus}{\textsc{Retro++}\xspace}

\definecolor{c0}{cmyk}{1,0.3968,0,0.2588} 
\definecolor{c1}{cmyk}{0,0.6175,0.8848,0.1490} 
\definecolor{c2}{cmyk}{0.1127,0.6690,0,0.4431} 
\definecolor{c3}{cmyk}{0.3081,0,0.7209,0.3255} 
\newtcbox{\hlprimary}{on line,colback=c0!10,colframe=white,size=fbox,arc=3pt, box align=base,before upper=\strut, top=-2pt, bottom=-4pt, left=-1pt, right=-1pt, boxrule=0pt}
\newtcbox{\hlprimarytab}{on line, box align=base, colback=c0!10,colframe=white,size=fbox,arc=3pt, before upper=\strut, top=-2pt, bottom=-4pt, left=-2pt, right=-2pt, boxrule=0pt}
\newtcbox{\hlsecondary}{on line,colback=c1!10,colframe=white,size=fbox,arc=3pt, box align=base,before upper=\strut, top=-2pt, bottom=-4pt, left=-1pt, right=-1pt, boxrule=0pt}
\newtcbox{\hlsecondarytab}{on line, box align=base, colback=c1!10,colframe=white,size=fbox,arc=3pt, before upper=\strut, top=-2pt, bottom=-4pt, left=-2pt, right=-2pt, boxrule=0pt}
\newtcolorbox{hlmultiline}{on line,colback=decentgrey!75,colframe=white,size=fbox,arc=3pt, box align=base, top=0pt, bottom=2pt, boxrule=0pt, before=\adjustbox{valign=c}\bgroup, after=\egroup, before upper=\strut}

\newcolumntype{Y}{>{\centering\arraybackslash}X}
\newcolumntype{Z}{>{\raggedleft\arraybackslash}X}

\newcommand{\dashifted}{{\tiny$\downarrow$}}
\newcommand{\da}[1]{{\scriptsize\hlprimarytab{\dashifted{#1}}}}

\newcommand{\uashifted}{{\tiny$\uparrow$}}
\newcommand{\ua}[1]{{\scriptsize\hlsecondarytab{\uashifted{#1}}}}

\definecolor{c4}{cmyk}{0.6765,0.2017,0,0.0667} 
\definecolor{c5}{cmyk}{0,0.8765,0.7099,0.3647} 

\definecolor{darkgrey}{RGB}{149,149,149}
\definecolor{decentgrey}{RGB}{242,242,242}

\title{Shall We Pretrain Autoregressive Language Models with Retrieval? \\ A Comprehensive Study}

\author{
Boxin Wang\thanks{\ \ Equal contribution. 
\ddag Work done during an internship at NVIDIA. $^1$UIUC. $^2$NVIDIA.  $^3$University of Wisconsin, Madison.
\quad \dag Correspondence to: Wei Ping <wping@nvidia.com>}\ \ $^\ddag$$^1$ 
  \And
  Wei Ping$^*$$^\dag$$^2$ 
  \And
  Peng Xu$^*$$^2$  
  \And
  Lawrence McAfee$^2$ 
  \AND
Zihan Liu$^2$
\And
 Mohammad Shoeybi$^2$
  \And
  Yi Dong$^2$
  \And
  Oleksii Kuchaiev$^2$
  \AND
  Bo Li$^1$
  \And
  Chaowei Xiao$^{2,}$$^3$
  \And
  Anima Anandkumar$^2$
  \And
  Bryan Catanzaro$^2$ 
}

\vspace{4cm}

\begin{document}
\maketitle
% \vspace{3cm}
\begin{abstract}
Large decoder-only language models~(LMs) can be largely improved in terms of perplexity by retrieval~(\textit{e.g.}, \method), but its impact on text generation quality and downstream task accuracy is unclear.
Thus, it is still an open question: \emph{shall we pretrain large autoregressive LMs with retrieval?}
To answer it, we perform a comprehensive study on a \emph{scalable pretrained} retrieval-augmented LM (i.e., \method) compared with standard GPT and retrieval-augmented GPT incorporated at fine-tuning or inference stages.  
We first provide the recipe to reproduce \method up to 9.5B parameters while retrieving a text corpus with 330B tokens. 
Based on that, we have the following novel findings:
\emph{i)} \method outperforms GPT on text generation with much less degeneration (i.e., repetition), moderately higher factual accuracy, and slightly lower toxicity with a nontoxic retrieval database.
\emph{ii)} On the LM Evaluation Harness benchmark, \method largely outperforms GPT on knowledge-intensive tasks, but is on par with GPT on other tasks.
Furthermore, we introduce a simple variant of the model, \methodplus, which largely improves open-domain QA results of original \method~(e.g., EM score $+8.6$ on Natural Question) and significantly outperforms retrieval-augmented GPT in both finetuning and zero-shot evaluation settings.
Our findings highlight the promising direction of pretraining autoregressive LMs with retrieval as future foundation models.
We release our code and model at: \url{https://github.com/NVIDIA/Megatron-LM/blob/main/tools/retro/README.md}.
\end{abstract}

\section{Introduction}
\label{sec:intro}
% \vspace{-.3em}
% 
Large language models~(LMs), including masked LMs~(e.g., BERT~\citep{devlin2018bert}), autoregressive LMs~(e.g., GPT~\citep{brown2020language}), and encoder-decoder LMs~(e.g., T5~\citep{raffel2020exploring}, BART~\citep{lewis2019bart}), have obtained state-of-the-art results for various NLP tasks.
Among them, the autoregressive LMs like GPT-3~\citep{brown2020language} and GPT-4~\citep{openai2023gpt4} demonstrate noticeable in-context learning ability and excellent long-form text generation results. 
Due to its importance, the community has spent considerable efforts to scale up such autoregressive generative LMs with more data and parameters and observed significant breakthroughs in a variety of real-world applications \citep[e.g.,][]{brown2020language}, including open-ended text generation and various downstream tasks~(e.g., question answering).
The successful public examples include GPT-3~(w/ 170B parameters)~\cite{brown2020language}, Gopher~(280B)~\citep{rae2021scaling}, Megatron-Turing~(530B)~\citep{smith2022using}, and PaLM~(540B)~\citep{chowdhery2022palm}.

Although large-scale autoregressive LMs have achieved huge successes, they also suffer from several weaknesses.
First, it requires a huge number of model parameters to memorize the world knowledge, which makes it costly for deployment.
Second, it is difficult to safeguard factual accuracy, which may provide users with incorrect information \citep{lee2022factuality}.
Third, it is expensive to update the model knowledge learned during pretraining with up-to-date facts~\citep{meng2022locating}, yielding outdated answers~\citep{lewis2020retrieval}. 

To mitigate the problems above, one line of research proposes to improve language models with retrieval.
The retrieval process can be integrated into LMs at:  \emph{i)} fine-tuning stage~\citep{karpukhin2020dense,lewis2020retrieval,guu2020retrieval}, or \emph{ii)} pretraining stage~\cite{borgeaud2022improving, izacard2022few}. 
 Most previous work augments {BERT or encoder-decoder LMs} with retrieval at fine-tuning stage, demonstrating successes for knowledge-intensive NLP tasks~\citep{guu2020retrieval,karpukhin2020dense,lewis2020retrieval, khandelwal2019generalization}. 
 {However, it remains relatively underexplored to pretrain \textit{autoregressive}~(decoder-only) LMs \textit{with retrieval}, especially considering the noticeable success of ChatGPT \citep{chatgpt} that underscores the extreme importance of the autoregressive LMs. }

 Most recently,  \method~\cite{borgeaud2022improving} proposes to pretrain autoregressive LMs with a retrieval module, which is practically scalable to large-scale pretraining from scratch by retrieving billions of token and largely reduces model parameters while achieving lower perplexity than standard GPT. 
It also provides the flexibility to update the knowledge stored in LMs~\cite{petroni2019language} by updating the retrieval database without training LMs again.
The success of pretraining LMs with retrieval raises an important question for the community if we want to pretrain autoregressive LMs  in the future: 
\emph{Shall we pretrain autoregressive~(decode-only) LMs with retrieval by default or not?}
However, previous work~\citep{borgeaud2022improving} misses the important evaluation on whether the model like \method could obtain comparable or even better results in terms of open-ended text generation and various NLP downstream tasks, apart from lower perplexity on the held-out dataset compared to standard GPT.

To answer the above \emph{question} and bridge the missing gap,  we perform an extensive study on \method, 
{as to the best of our knowledge, \method is the only retrieval-augmented autoregressive LM that supports large-scale pretraining with retrieval on the massive pretraining corpus with hundreds of billion or trillion tokens.}
Our comprehensive study sheds light on the promising direction of pretraining autoregressive LMs with retrieval to serve as future foundation models, as they overall outperform standard GPT models in terms of perplexity, text generation quality, and downstream task performances, especially for knowledge-intensive tasks, including open-domain QA. 

% \vspace{-.2em}
\section{Key Findings}
% \vspace{-.2em}

\begin{table*}[t]\small
\centering
\resizebox{1.0\linewidth}{!}
{
\begin{tabular}{lccccc}
\toprule
{Model} & {\#/ Retrieval} & {When to} & \multirow{2}{*}{Architecture} & \multirow{2}{*}{Initialization} & \multirow{2}{*}{Re-indexing} \\
Name                            &       Tokens                                & Involve Retrieval        &                           &                              \\

\midrule
\method~(\citeauthor{borgeaud2022improving}) & $O(10^{12})$ & Pretraining & decoder-only & From Scratch / Pretrained GPT & No \\
Atlas~(\citeauthor{izacard2022few}) & $O(10^{9})$ & Pretraining & encoder-decoder & Pretrained T5 & Yes \\
REALM~(\citeauthor{guu2020retrieval}) & $O(10^{9})$ & Pretraining & encoder-only & Pretrained BERT & Yes \\
\midrule
RAG~(\citeauthor{lewis2020retrieval}) & $O(10^{9})$ & Fine-tuning & encoder-decoder & Pretrained BART & No \\
DPR~(\citeauthor{karpukhin2020dense}) & $O(10^{9})$ & Fine-tuning & encoder-only & Pretrained BERT & No \\
FiD~(\citeauthor{izacard2021leveraging}) & $O(10^{9})$ & Fine-tuning & encoder-decoder & Pretrained T5 & No \\
% WebGPT~(\citeauthor{nakano2021webgpt}) & Fine-tuning & decoder-only & Pretrained GPT & No \\
\midrule
KNN-LM~(\citeauthor{khandelwal2019generalization}) & $O(10^{9})$ & Inference & decoder-only & Pretrained GPT & No \\
\bottomrule
\end{tabular}
}
% \vspace{-2mm}
\caption{{\small Comparison of different retrieval-augmented models in terms of \#/ retrieval tokens, which stage to incorporate retrieval into LMs, the architecture of the backbone LM, whether it requires initialization from the existing LM checkpoint, and whether it requires expensive re-indexing. \method is the most scalable retrieval-augmented LM due to its chunk-level retrieval and scalable decoder-only autoregressive LM backbone \citep{thoppilan2022lamda,brown2020language,smith2022using,chowdhery2022palm} without expensive retrieval index refresh.}}
\label{table:comparison}
% \vspace{-2mm}
\end{table*}

We successfully reproduce and pretrain \method\citep{borgeaud2022improving}  from scratch\footnote{The official implementation and pretrained checkpoints are not open-sourced.},
with parameter sizes ranging from 148M up to 9.5B by retrieving from a text corpus with over 330B tokens. 
In addition, we discuss the inference strategy of \method for text generation that is not covered in \citet{borgeaud2022improving}, and perform a large-scale evaluation in different scenarios.

To minimize the discrepancy variables between \method and GPT,  we use the same decoder architecture, same hyper-parameters, and same pre-training corpus to pre-train \method and GPT given the same number of pre-training steps. We highlight our novel findings for \method and GPT as follows:

\subsection{Text Generation}
We conduct a systematic study (see \S\ref{sec:generation}) to understand and analyze \method by evaluating its open-ended text generation quality via human and automatic evaluations.
\method exhibits better performance than GPT with considerably less \emph{repetition}, moderately higher \emph{factual accuracy}, and slightly lower \emph{toxicity} levels. \method is on par with GPT in terms of \emph{fluency}, \emph{coherence}.

\subsection{LM Evaluation Harness Benchmark}
 In terms of zero-shot evaluation on the standard benchmark, \method can overall improve upon the GPT across different tasks, significantly outperforming GPT on knowledge-intensive tasks such as Hellaswag and BoolQ while achieving similar performance on other tasks.
 Specifically, we evaluate the zero-shot capabilities of \method and GPT on nine representative NLP downstream classification tasks (see \S\ref{sec:downstream}). 
{Additionally, our findings demonstrate that \method can leverage retrieved neighbors and significantly improves accuracy for knowledge-intensive tasks in zero-shot evaluations. In contrast, incorporating these retrieved neighbors directly during the inference stage can hurt GPT's performance. These results further substantiate the potential of \method, which is pre-trained with retrieval capabilities, as a promising approach.}

\subsection{Open-domain QA}
For open-domain QA tasks, \method achieves considerably superior performance than retrieval-augmented GPT that incorporates retrieval during fine-tuning across different model sizes and datasets.  
Specifically, we propose a variant of the model, \methodplus, for open-domain QA that feeds the most relevant evidence into the decoder and more evidence into its encoder, which is different from the original version~\citep{borgeaud2022improving}. \methodplus can largely improve the exact matching score~(EM) on Natrual Question from 40.9\% to 54.1\%, which is significant higher than the 45.5\% reported by the original \method.

% \vspace{-.3em}
\section{Related Work}
% \vspace{-.3em}
\label{sec:related}
Retrieval has been applied in various NLP tasks for years, including question answering~(QA)~\citep[e.g.,][]{bilotti2007structured},  machine translation~\citep[e.g.,][]{zhang2018guiding}, and conversation~\citep{shuster2021retrieval,thoppilan2022lamda, komeili2021internet}.
In particular, language models have been augmented with retrieval at different stages, including inference time~\citep{khandelwal2019generalization,yogatama2021adaptive}, fine-tuning stage~\citep{karpukhin2020dense, lewis2020retrieval, guu2020retrieval}, and pretraining stage~\cite{borgeaud2022improving, izacard2022few}.
LMs have been augmented with retrieval at the fine-tuning stage for downstream tasks, primarily for open-domain QA.
DPR~\citep{karpukhin2020dense} finetunes one BERT to encode questions and the other BERT to encode answers within a dual encoder framework, using a contrastive loss to align the hidden representations of question and corresponding answer.
RAG~\citep{lewis2020retrieval} studies the fine-tuning recipe for retrieval-augmented generation models, especially on open-domain QA tasks.
{FiD~\citep{izacard2021leveraging} improves RAG with a better LM backbone T5, and fuses multiple retrieved passages to the decoder during fine-tuning to further improve QA accuracy.
WebGPT~\citep{nakano2021webgpt} leverages web search engine and fine-tunes GPT using reinforcement learning with human feedback (RLHF) for reference generation and factuality improvement, which is orthogonal to our work that focuses on pretraining with retrieval. The proposed RLHF can be applied to \method as well. 
}

REALM~\citep{guu2020retrieval} performs both unsupervised pretraining and supervised fine-tuning strategies for retrieval-augmented BERT model in open-domain QA. 
Their pretraining involves asynchronous re-embedding and re-indexing all documents every several hundred training steps, which quickly becomes impractical for training corpus with trillion tokens.
Atlas~\citep{izacard2022few} uses a similar approach but augments the T5 architecture~\citep{raffel2020exploring} with retrieval at both pre-training and fine-tuning. Before pretraining, it first initializes the encoder-decoder LM backbone with pretrained T5, and the dense retriever with pretrained Contriever~\citep{izacardunsupervised}. During pretraining, it also applies asynchronous index refresh every 1000 steps.

In contrast, \method~\citep{borgeaud2022improving} embeds and indexes the whole training corpus at chunk-level~(e.g., chuck size = 64) with a frozen BERT before pretraining.
During pretraining, the model relies on a trainable bidirectional encoder to embed the retrieved chunks of raw text. The GPT decoder further ``select'' the relevant piece of evidence from the encoder side by a chunk-wise cross-attention. This architecture design enables  LM pretraining on hundreds of billion tokens by retrieving from trillion tokens. 
See Table~\ref{table:comparison} for a complete comparison of retrieval-augmented LMs.

\begin{figure*}[t]
\begin{subfigure}{.33\textwidth}
    \centering
    \includegraphics[height=0.67\linewidth]{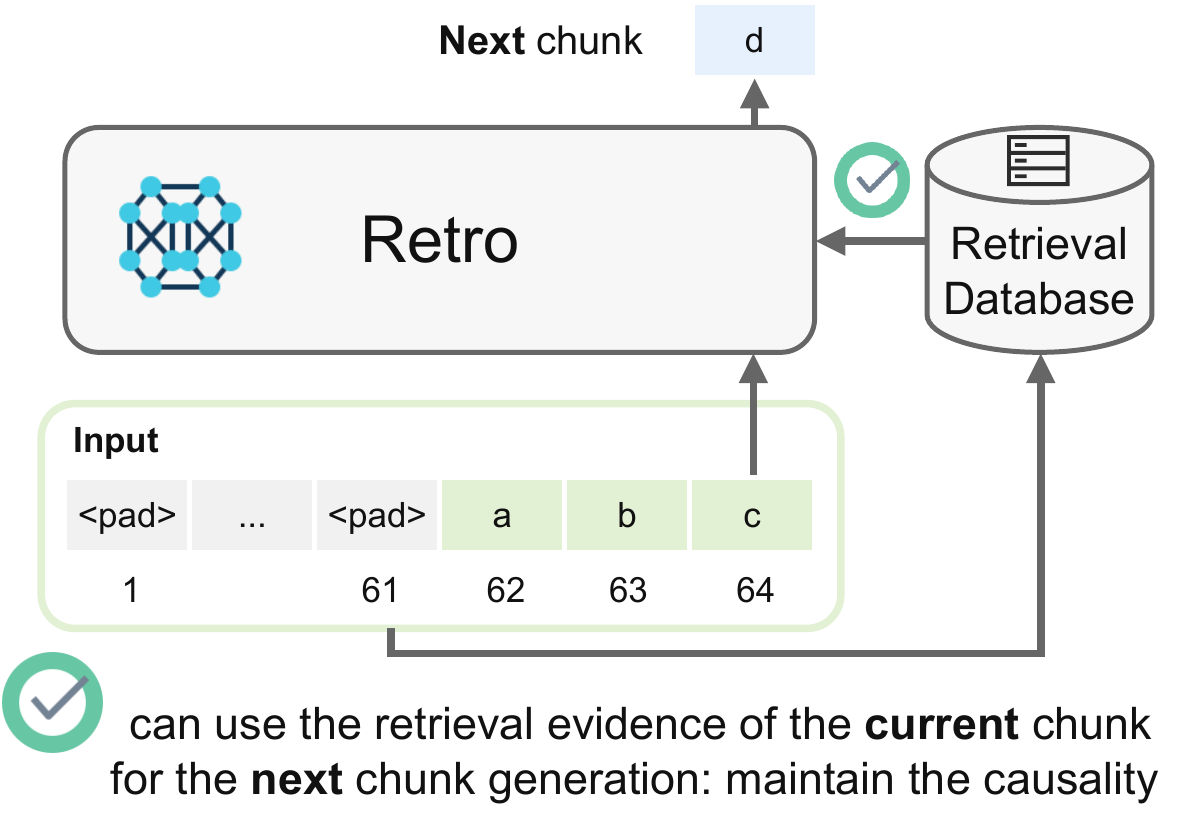}
    \caption{Use ``left padding'' Rule}
    \label{fig:2}
\end{subfigure}
\begin{subfigure}{.33\textwidth}
    \centering
    \includegraphics[height=0.67\linewidth]{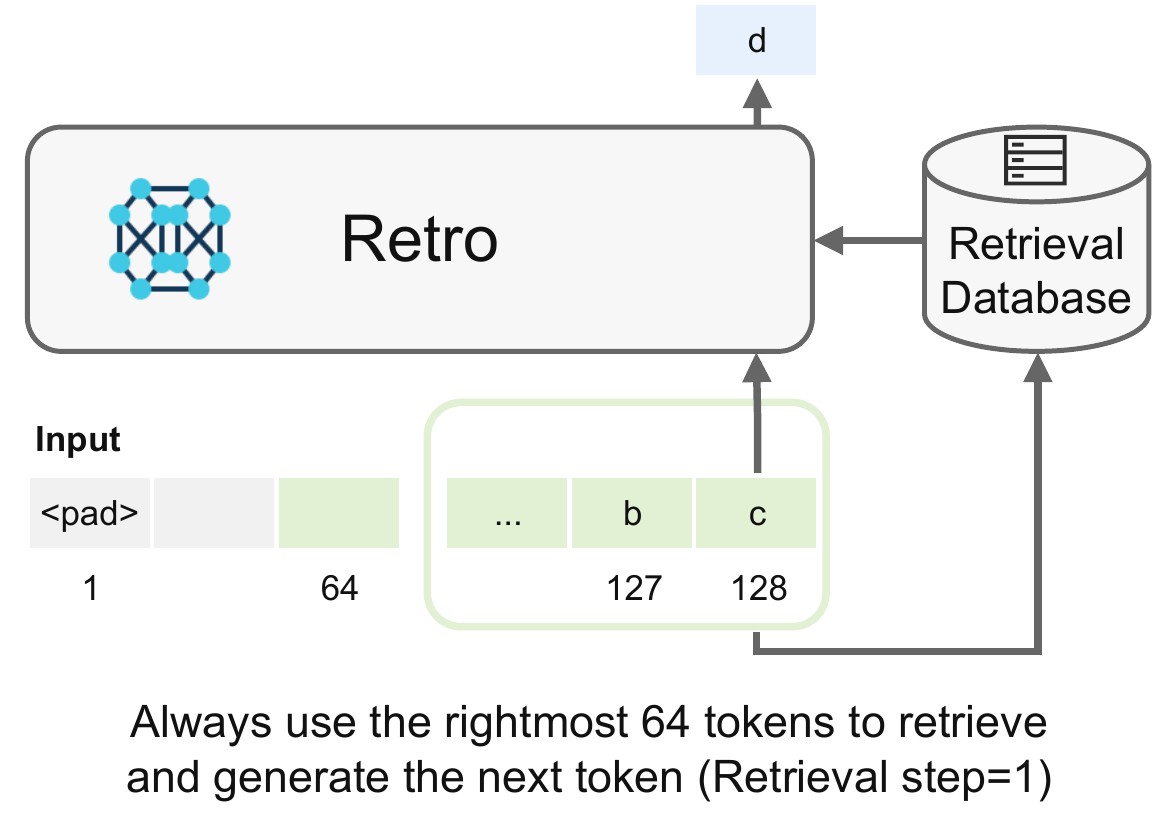}
    \caption{Retrieval step $=1$}
    \label{fig:4}
\end{subfigure}
\begin{subfigure}{.33\textwidth}
    \centering
    \includegraphics[height=0.67\linewidth]{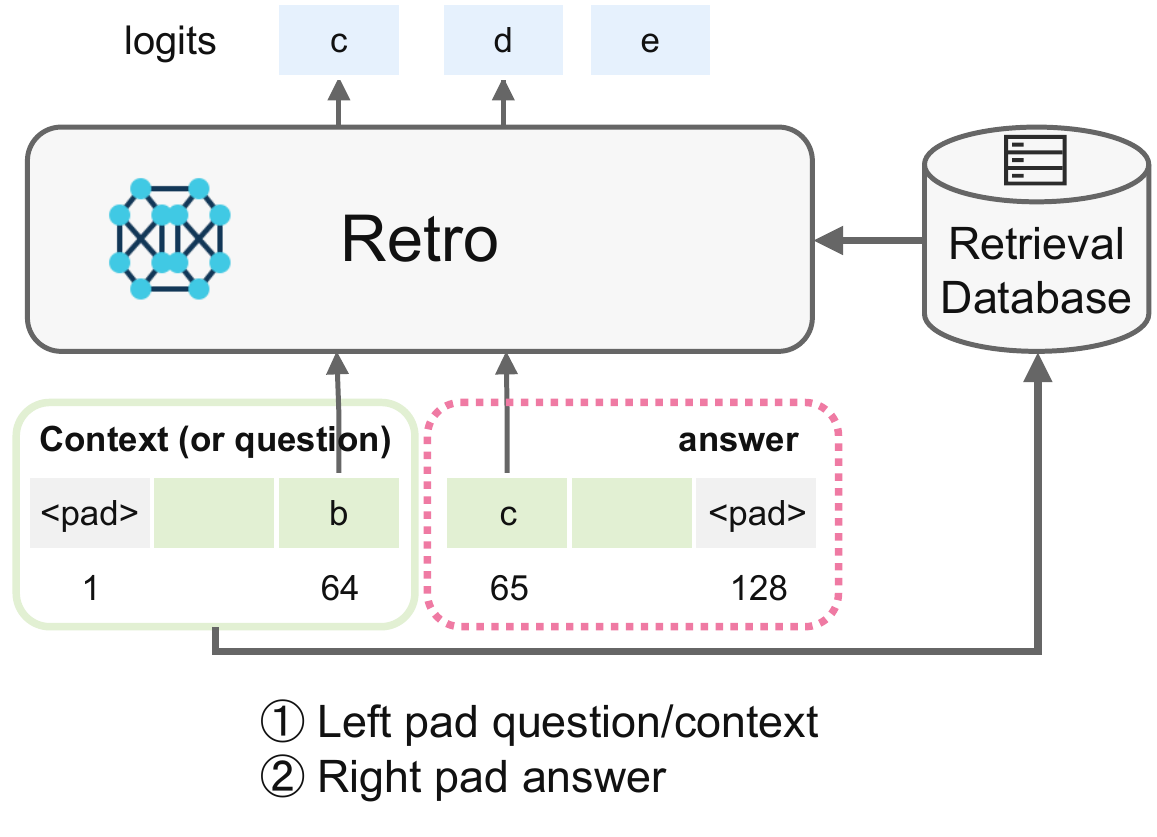}
    \caption{Separate question and answer chunks}
    \label{fig:6}
\end{subfigure}
\caption{Visualization of padding design for \method.}
% \vspace{-3mm}
\end{figure*}

% \vspace{-.1em}
\section{Model and Implementation}
\label{sec:method}
% \vspace{-.3em}
In this section, we first introduce preliminaries of \method, then provide detailed recipe of our implementation, including retrieval database, pretraining, and retrieval-augmented finetuning and generation.  

\begin{table}[t] \small
\centering
\begin{tabular}{l|llll}
\toprule
                       & Small  & Medium & XL & XXL \\
\midrule
{GPT}           & 17.76 & 13.18 & 10.18  & 7.86  \\
\method($k=2$)         &  12.99     &  10.06     &   8.10     &  6.72    \\
\bottomrule
\end{tabular}
\vspace{-.4em}
\caption{\small Validation perplexity of pretrained GPT and \method on the held-out  dataset. We report the results with $k=2$ neighbors in this Table, and we observe the same trend of improvements with larger $k$ as in \citet{borgeaud2022improving}.}
\label{tab:ppl}
\vspace{-2mm}
\end{table}

% \vspace{-.3em}
\subsection{Preliminaries of \method}
% \vspace{-.2em}
% \noindent \textbf{Chunk-wise retrieval-augmented LM.}
\method is an autoregressive language model enhanced with a retrieval module that utilizes chunk-wise retrieval, enabling it to scale up to trillions of tokens. 
The model splits both the input sequence and retrieval datastore into sequences of chunks. \textsc{Retro} retrieves nearest neighbor chunks from the retrieval database using the previous chunk and fuses this information with the context from preceding chunks to guide the generation of the next chunk. To maintain causality, the model can only use the nearest neighbors of the previous chunk for the autoregressive generation. 
% chunk
% retrieval neighbor causal relationship

% \vspace{-.3em}
\subsection{Implementation}
% \vspace{-.2em}
As \method has no official open-source implementation and pretrained checkpoints, we reproduce and pretrain \method from scratch on our own.

% \vspace{-.3em}
\subsubsection{Retrieval Database}
We build the retrieval database with the whole pretraining dataset mentioned in \S\ref{sec:gpt}. In this way, \method and standard GPT of similar size are fair comparisons, as they are pretrained using the same information from the pretraining corpus. 
The retrieval database is a key-value database, where values are chunks split from the pretraining corpus, and the keys are corresponding BERT embeddings. 
Our pretraining dataset with 330B tokens yields a retrieval database consisting of 5.3B chunks in total with chunk size $m=64$.

\noindent \textbf{Retrieval Index.} 
We use the Faiss index \citep{faiss} as the implementation for the dense retriever to search for approximate nearest neighbors in the BERT embedding space.
We configure the Faiss index to cluster the dense embeddings into $2^{22}$ centroids accelerated with Hierarchical Navigable Small World graphs \citep{hnsw} to speed up the query. 
We also encode the embeddings with optimized product quantization \citep{pq,opq} to compress memory overhead and further improve the query throughput.
As a result, we can achieve 4\textit{ms} per query over the whole pretraining corpus averaged for each chunk on a DGX-2H node.
One may find more details in Appendix~\S\ref{app:faiss}.

\subsubsection{Pretraining \method Models} 
We use the same transformer configurations (\#/ layers, hidden size, attention heads) and pretrain both \method and standard GPT from scratch. 
Specifically, we pretrain \method across different parameter sizes, ranging from 148M (Small),  410M (Medium), 1.5B (XL), and 9.5B (XXL). 
We also use the same pretraining schedules to pretrain \method and GPT given the same number of steps.
We list the validation perplexity of GPT and \method after pretraining in Table \ref{tab:ppl}. 
We present more details in Appendix \S\ref{app:lm}, including pretraining schedules, computational cost~(GPU hours), and model architectures.

\begin{table*}[tbh!]\small
% \vspace{-5mm}
    \centering
    % \resizebox{\linewidth}{!}
    {
    \begin{tabular}{l|lc|cc|cc|cc}
    \toprule
\multicolumn{1}{l|}{\multirow{2}{*}{\textbf{Metrics}}} & \multicolumn{2}{c|}{\textbf{Small}}  & \multicolumn{2}{c|}{\textbf{Medium}}  & \multicolumn{2}{c|}{\textbf{XL}}  & \multicolumn{2}{c}{\textbf{XXL}}  \\
& GPT & \method & GPT & \method   & GPT & \method  & GPT & \method   \\ 
\midrule
Repetition \%    & $2.86\%$ & $\textbf{2.26}\%$   & $1.70\%$ & $\textbf{1.50}\%$ & $1.44\%$ & $\textbf{0.96}\%$  & $1.40\%$ & $\textbf{1.12}\%$    \\
Self-BLEU       & $0.29$   & $0.3$      & $0.29$   & $0.3$    & $0.29$   & $0.29$    & $0.31$   & $0.31$  \\
Zipf Coefficient & $0.98$   & $0.98$     & $0.96$   & $0.98$   & $0.97$   & $0.98$    & $0.96$   & $0.96$\\
\bottomrule
\end{tabular}
}
% \vspace{-2mm}
\caption{\small Automatic evaluation on text generation quality for \method and GPT across different sizes. }
\label{tab:quality}
% \vspace{-3mm}
\end{table*}

% \vspace{-.3em}
\subsubsection{Retrieval-augmented Generation}
% \vspace{-.2em}
We discuss the generation and inference recipe in the batch-processing mode for \method, which is missing from the previous literature.

\noindent \textbf{``Left Padding'' Rule.} The chunk-wise retrieval of \method improves scalability but enforces chunk-wise alignment constraints, leading to issues in conditional generations with short contexts. When the sequence length is less than the chunk size, \method cannot utilize its retrieval capability as there is no previous chunk for retrieval. 
Instead, \method adds padding tokens to the left of the context, allowing \method to leverage the retrieved neighbors from the previous context to guide the generation of the next token (\Cref{fig:2}).
We summarize this general principle in \method as the ``left padding'' rule, as it can leverage the contextual information for retrieval to the most.
This rule remains preferable for input sequences larger than the chunk size, as it ensures the closest and rightmost context is used for retrieval, making it more relevant for next token prediction (see \Cref{fig:4}).

\noindent \textbf{Frequency of Retrieval.}
In order to efficiently generate long sequences with \method, we note a flexible trade-off between retrieval-augmented generation and computation overhead. 
The direct method involves retrieval at every decoding step, maximizing the use of the retrieval module but increasing computational overhead (\Cref{fig:4}, retrieval step $=1$). 
Another approach retrieves neighbors at the frequency of the chunk size, reducing overhead but sacrificing accuracy (Appendix \Cref{fig:3}, retrieval step $=64$).
To balance these factors, we introduce a flexible retrieval step, which allows model practitioners to choose how many tokens to generate with the current retrieved neighbors before updating the context. 
Smaller retrieval steps are preferred for downstream tasks with short answers to ensure accurate neighbors, while larger steps are used for efficient generation of long passages. We provide more details in  Appendix \S\ref{app:generation}.

\subsubsection{Batched Training for Downstream Tasks}
When fine-tuning \method for downstream tasks~(e.g., QA), it is crucial to separate context or question from the candidate answer chunk to maintain causality in autoregressive modeling. 
This leads to a modified "left padding" rule: pad \textit{context chunks} from the left and \textit{answer chunks} from the right (\Cref{fig:6}). Padding aligns input sequences with the chunk size, enabling batch-mode training and inference for faster evaluation. By adding padding chunks to the right, sequences with varying chunk numbers can be processed together, further improving efficiency.

\section{Open-ended Text Generation}
\label{sec:generation}
In this section, we delve into the problem of open-ended text generation, which refers to tasks of generating coherent continuation given the preceding prompt.
Given that this problem for \method has never been studied before,
we manage to bridge the gap and evaluate the open-ended text generation of \method compared to GPT from three aspects: $a$) text quality, $b$) factuality, and $c$) toxicity.

\subsection{Text Quality}
We perform both automatic and human evaluations.
\subsubsection{Automatic Evaluation}

\begin{table*}[!ht]\small
% \vspace{-5mm}    
   \begin{subtable}{.5\linewidth}
    \centering
    \resizebox{\linewidth}{!}
    {
    \begin{tabular}{cc|cc|cc}
    \toprule
        \multirow{2}{*}{\textbf{Decoding}} & \multirow{2}{*}{\textbf{Models}} &  \multicolumn{2}{c|}{\textbf{Factual}}  & \multicolumn{2}{c}{\textbf{Nonfactual}}   \\ 
        & ~ & $\text{NE}_\text{ER}\downarrow$ & $\text{Entail}_\text{R}\uparrow$ & $\text{NE}_\text{ER}\downarrow$ & $\text{Entail}_\text{R}\uparrow$  \\ \midrule
       \multirow{2}{*}{\textit{Top-p=0.9}} & \method & \textbf{52.14\%} & \textbf{3.11\%} & \textbf{56.75\%} & \textbf{2.06\%}  \\ 
        ~ & GPT & 52.42\% & 2.93\% & 56.82\% & 2.04\%  \\ \midrule
        % \textit{Factual-} & \method  & \textbf{40.80\%} & \textbf{10.73\%} & {46.66\%} & \textbf{6.17\%}  \\
        % \textit{Nucleus} & GPT & 42.10\% & 10.10\% & \textbf{46.50\%} & 5.60\%  \\  \midrule
        \multirow{2}{*}{\textit{Greedy}} & \method  & \textbf{37.42\%} & \textbf{16.66\%} & \textbf{42.45\%} & \textbf{10.88\%}  \\ 
        ~ & GPT & 39.87\% & 12.91\% & 45.02\% & 8.75\% \\ 
        \bottomrule
    \end{tabular}
    }
   \caption{\small The factuality on \textsc{FactualityPrompts} benchmark.}
   \label{tab:factual}
    \end{subtable}
    \begin{subtable}{.5\linewidth}
    \centering \small
    \resizebox{0.93\linewidth}{!}
    {
    \begin{tabular}{c|cc|cc}
    \toprule
        \multirow{2}{*}{\textbf{Models}} &  \multicolumn{2}{c|}{\textbf{QA Format}}  & \multicolumn{2}{c}{\textbf{Null Format}}   \\ 
         ~ & MC1$\uparrow$ & MC2$\uparrow$  & MC1$\uparrow$ & MC2$\uparrow$   \\ \midrule
        GPT                 & $0.222$ & $0.377$ & $0.234$ & $0.435$ \\
        \method (pretraining) & $\textbf{0.239}$ & $\textbf{0.382}$ & $\textbf{0.248}$ & $\textbf{0.439}$ \\
        \method  (wiki)        &   -     &  -      & $0.242$ & $0.437$\\
        \method  (DPR)         &   -     &  -      & $0.245$ & $\textbf{0.439}$\\ \bottomrule
    \end{tabular}
    }
    \caption{\small The truthfulness on TruthfulQA benchmark.}
      \label{tab:truthful}
    \end{subtable}
% \vspace{-4mm}    
    \caption{\small Evaluation of factuality and truthfulness of \method~(XL) and GPT~(XL).}
% \vspace{-2mm}
\end{table*}

\noindent \textbf{Evaluation Metrics.} We follow prior work \citep{nucleus,selfbleu} and  consider the following metrics: \textbf{Repetition \%} measures percentage of the generations containing repetitive phrases, \textbf{SELF-BLUE} evaluates the diversity of the generations, and \textbf{Zipf Coefficient} measures the use of vocabulary. See detailed definition and evaluation setup in Appendix \S\ref{app:auto-eval}.

\noindent \textbf{Experimental Results.} Our results are shown in Table \ref{tab:quality}.
We note that \method can reduce the percentage of repetition compared with GPT by a large margin across different sizes.
Specifically, \method averagely mitigates 21\% of repetitions compared with GPT across different sizes.
This suggests the retrieval module can help reduce text degeneration by referencing retrieved human text. 
Regarding vocabulary use and generation diversity, we do not observe major differences between GPT and \method, which implies these properties are primarily dependent on the decoder component of LMs.

\subsubsection{Human Evaluation}
We also conduct human evaluations to further verify the quality of the generated text.

\noindent\textbf{Evaluation Metrics.} We ask human annotators to annotate each generation with fluency scores, which measure the human readability and grammatical errors from 1 (Not human-readable) to 5 (Very fluent), and coherence scores, which measure the relevance between the prompt and the corresponding continuations from 1 (Not Relevant) to 5 (Very Relevant). More details can be found in \S\ref{app:human-eval}.

\noindent\textbf{Experimental Results.} We present the human vote histogram in Appendix Figure \ref{fig:votes}.
We observe that most votes concentrate on the regime of scores $>= 3$ for both relevance and fluency, which indicates that our generated text from both models is of high quality and closely related to the prompts. 
The differences between GPT and \method are subtle, with average relevance (3.726) and fluency (3.826) scores of \method slightly outperforming the average relevance score (3.715) and fluency (3.818) scores of GPT.

From both automatic and human evaluation, we can conclude that although the generation of \method adds some complexity, we do not see any sign of the degeneration of \method compared to GPT. Moreover, \method is shown to be able to reduce the repetition and slightly improve text generation quality.

\subsection{Factuality}

Factuality refers to being coherent to provide ground truth knowledge sources in NLP. 
We leverage two well-established benchmarks \citep{lee2022factuality,lin2021truthfulqa} to evaluate the factual accuracy of \method and GPT. We put detailed experimental setup and metrics in the \S\ref{sec:appendfact}.

\noindent \textbf{Factuality Prompts.}
We present the results on \textsc{FactualityPrompts} benchmark in Table \ref{tab:factual}, where we follow the protocol~\cite{lee2022factuality} and use different decoding strategies for evaluation. 
We observe that overall \method demonstrates better factual accuracy and makes fewer hallucinations. 
Our observation also aligns with \citep{lee2022factuality} that different from top-$p$ sampling with randomness, greedy decoding gives the much better factual accuracy by sacrificing sample diversity.
\method shows a significant improvement over GPT using both top-p and greedy decoding algorithms.

\noindent \textbf{TruthfulQA.} We also evaluate truthfulness in Table \ref{tab:truthful}, where we observe the consistent conclusion that \method tend to be more resilient to human falsehoods than GPT across different prompts (demonstrated in Appendix Table \ref{tab:qa-prompt} and \ref{tab:null-prompt}).

Since using the null format yields consistently better performance, we thus fix the prompt and conduct ablation studies with different retrieval databases. 
We observe that the pretraining database and DPR database yield the best factual accuracy.
This suggests that ($i$) finetuned retriever or ($ii$) retrieval from a larger corpus are helpful to improve factual accuracy.
In addition, the improvement margin for \method on TruthfulQA is not as significant as on \textsc{FactualityPrompts}.
This is likely because TruthfulQA is an adversarial dataset containing difficult question types, such as logical falsehoods and indexical errors (e.g., ``Are you an artificial intelligence?'', more examples can be found in Appendix \Cref{tab:truthful-examples}). In these cases, retrieval from the database does not effectively aid in answering such questions.

\begin{table*}[t]\small
    \label{tab:standard}
        \centering
        % \begin{adjustbox}{width=\textwidth}
        % \resizebox{\linewidth}{!}
        {
        \begin{tabular}{l|c|lcc|lcc}
        \toprule
    \multicolumn{1}{c|}{\multirow{2}{*}{\textbf{Models}}} & \textbf{Retrieval}  &  \multicolumn{3}{c|}{\textbf{Exp. Max. Toxicity} ~($\downarrow$)} &  \multicolumn{3}{c}{\textbf{Toxicity Prob.} ~($\downarrow$)}  
    \\
   & \textbf{Database} & \textbf{Full} & \textbf{Toxic} & \textbf{Nontoxic} & \textbf{Full} & \textbf{Toxic} & \textbf{Nontoxic}  \\
    \midrule
    GPT & - & $0.44$ & $0.64$ & $0.39$  & 37\% & 74\% & 27\%  \\ 
    \midrule
    \method (top-$N=2$, top-$K=2$) & Pretraining & $0.46$ & $0.66$ & $0.40$  & 40\% & 76\% & 30\%    \\
     \method  (top-$N=5$, top-$K=2$) & Pretraining  & $0.46$ & $0.66$ & $0.40$  & 39\% & 77\% & 29\%    \\
     \method  (top-$N=10$, top-$K=2$) & Pretraining & $0.46$ & $0.66$ & $0.40$  & 39\% & 76\% & 29\%    \\
    \midrule
     \method  (top-$N=2$, top-$K=2$) & Wiki  & $0.43$ & $0.64$ & $0.38$  & 35\% & 73\% & 25\%    \\
     \method  (top-$N=5$, top-$K=2$) & Wiki & $0.43$ & $0.64$ & $0.38$  & 35\% & 71\% & 26\%    \\
     \method  (top-$N=10$, top-$K=2$) & Wiki & $0.43$ & $0.64$ & $0.38$  & 35\% & 71\% & 26\%    \\
    \bottomrule 
    \end{tabular}
    }
        % \vspace{-.1cm}
    \caption{\small Evaluation of LM toxicity for GPT (XL) and \method (XL).
    Model toxicity is evaluated on \textsc{RealToxicityPrompts}.
    \textbf{Full} refers to the full set of prompts, \textbf{Toxic} and \textbf{Nontoxic} refer to the toxic and nontoxic subsets of prompts.
    $\downarrow$ means the lower, the better. 
    \method can filter from top-$N$ nearest neighbors and select the top-$K$ nontoxic neighbors for retrieval.
    }
    \label{tab:toxicity}
    % \vspace{-1mm}
\end{table*}

\subsection{Toxicity}

The toxicity of LMs refers to the possibility of LMs that output toxic generations. In this study, we follow \textsc{RealToxictyPrompts} benchmark \citep{gehman2020realtoxicityprompts} to evaluate the potential toxicity of \method and GPT.

\noindent \textbf{Evaluation Metrics.} 
Following \citet{gehman2020realtoxicityprompts}, we report the \emph{Expected Maximum Toxicity}, which evaluates the toxicity of the worst-case generation, as well as \emph{Toxicity Probability} that estimates the empirical frequency of generating toxic language. See more details and setup in \S\ref{app:toxicity-setup}.

\noindent  \textbf{Experimental Results.} 
The toxicity of LMs are shown in Table \ref{tab:toxicity}. 
Compared to GPT, we note that \method with the pretraining corpus even increases the toxicity of the generations. 
Moreover, we observe more toxicity increases in toxic prompts than in nontoxic prompts. This suggests that when prompting \method with toxic contexts, it is more likely to retrieve toxic evidence and thus amplify the issues. 
To confirm the toxicity amplification issue, we further conduct two sets of ablation studies:
($i$) We save the retrieval evidence and calculate the \textit{Expected \textbf{Mean} Toxicity} of both generations and retrieval evidence. We observe that the toxicity of retrieval evidence is $0.177$, higher than the toxicity of the generations ($0.146$).
($ii$) We change the retrieval database to the Wikipedia database, which shows lower toxicity for retrieval evidence ($0.132$). As a result, we observe that \method with the Wikipedia retrieval database can help mitigate the toxicity of GPT as shown in Table \ref{tab:toxicity}, with the toxicity probability dropping from $37\%$ to $35\%$. 
We also note that it is not very helpful to use a larger $N$ as nearest neighbors and filter the retrieval evidence by toxicity. We hypothesize the reason is that the similarity between input and retrieval evidence is limited with larger $N$, thus yielding low cross-attention on the retrieval evidence.

\begin{table*}[htp!]\small
    \centering
    % \resizebox{\linewidth}{!}
    {
    \begin{tabular}{l|lc|cc|cc|cc}
    \toprule
\multicolumn{1}{l|}{\multirow{2}{*}{\textbf{Tasks}}} & \multicolumn{2}{c|}{\textbf{Small}}  & \multicolumn{2}{c|}{\textbf{Medium}}  & \multicolumn{2}{c|}{\textbf{XL}}  & \multicolumn{2}{c}{\textbf{XXL}}    \\
& GPT & \method & GPT & \method   & GPT & \method  & GPT & \method   \\ 
\midrule
\multicolumn{1}{l}{\textit{Knowledge-intensive Tasks}} \\
\midrule
HellaSwag  & $31.3$ & $36.2$ \ua{4.9}  & $43.2$ & $46.2$ \ua{3.0}  & $56.7$ & $59.0$ \ua{2.3}  & $72.3$ & $70.6$ \da{1.7}  \\
BoolQ      & $59.3$ & $61.8$ \ua{2.5}  & $57.4$ & $57.2$ \da{0.2}  & $62.2$ & $62.7$ \ua{0.5}  & $67.3$ & $70.7$ \ua{3.4}   \\
\midrule
\multicolumn{1}{l}{\textit{Knowledge-nonintensive Tasks}} \\
\midrule
Lambada    & $41.7$ & $41.4$ \da{0.3}  & $54.1$ & $55.0$ \ua{0.9}  & $63.9$ & $64.0$ \ua{0.1}  & $73.9$ & $72.7$ \da{1.2}    \\
RACE       & $34.6$ & $32.5$ \da{2.1}  & $37.3$ & $37.3$ \ua{0.0}  & $40.8$ & $39.9$ \da{0.9}  & $44.3$ & $43.2$ \da{1.1}  \\
PiQA       & $64.3$ & $64.8$ \ua{0.5}  & $70.2$ & $68.7$ \da{1.5}  & $73.7$ & $74.1$ \ua{0.4}  & $78.5$ & $77.4$ \da{1.1}  \\
WinoGrande & $52.4$ & $52.0$ \da{0.4}  & $53.8$ & $55.2$ \ua{1.4}  & $59.0$ & $60.1$ \ua{1.1}  & $68.5$ & $65.8$ \da{2.7}   \\
ANLI-R2    & $35.1$ & $36.2$ \ua{1.1}  & $33.5$ & $33.3$ \da{0.2}  & $34.3$ & $35.3$ \ua{1.0}  & $32.2$ & $35.5$ \ua{3.3}    \\
HANS       & $51.5$ & $51.4$ \da{0.1}  & $50.5$ & $50.5$ \ua{0.0}  & $50.1$ & $50.0$ \da{0.1}  & $50.8$ & $56.5$ \ua{5.7}   \\
WiC        & $50.0$ & $50.0$ \ua{0.0}  & $50.2$ & $50.0$ \da{0.2}  & $47.8$ & $49.8$ \ua{2.0}  & $52.4$ & $52.4$ \ua{0.0}  \\
\midrule     
Avg.  Acc. ($\uparrow$)  & $46.7$  & $47.4$ \ua{0.7}  & $50.0$  & $50.4$ \ua{0.4} & $54.3$  & $55.0$ \ua{0.7} & $60.0$ &  $60.5$ \ua{0.5}   \\
\bottomrule
\end{tabular}
}
% \vspace{-1mm}
\caption{\small Accuracy (Acc.) on nine downstream tasks evaluated in the zero-shot setting for pretrained LMs with different parameter sizes. 
}
\label{tab:zeroshot-full}
% \vspace{-1mm}
\end{table*}

% \vspace{-.3em}
\section{LM Evaluation Harness Benchmark}
\label{sec:downstream}
% 9 tasks
% \vspace{-.3em}
Besides the open-ended text generation, it is also important to examine the generalization of \method on various downstream tasks, which is also missing from the literature. 
Therefore, we use LM Evaluation Harness Benchmark \citep{lmharness} and consider the following nine representative NLP downstream tasks.
See more details in \S\ref{app:harness}.

\noindent \textbf{Zero-shot evaluation.} 
We present the zero-shot evaluation results in Table \ref{tab:zeroshot-full}. We find that on average \method can improve the downstream task accuracy across different tasks.
Moreover, we observe larger improvements in knowledge-intensive tasks such as Hellaswag and BoolQ~(6 of 8 cases), which require factual knowledge to guide the reasoning. Note that the zero-shot evaluation results are susceptible to prompt formats, so the results have certain variances.

\noindent  \textbf{Retrieval-augmented GPT at Inference time.}
We have seen that retrieval significantly improves \method across different downstream tasks in the zero-shot setting. 
In this ablation study, we append the retrieval evidence of \method to the beginning of the context to see whether retrieval can also be helpful for GPT at inference time. 
We evaluate the zero-shot accuracy after prepending the top-$1$ retrieval evidence. The results are shown in Appendix Table \ref{tab:zeroshot-hellaswag}. We observe that directly prepending the evidence from the retrieval database messes up the GPT context in the zero-shot setting, yielding low accuracy of around $24.5\%$. We hypothesize the reason is that the retrieval evidence can be noisy. Without pretraining or proper fine-tuning, GPT in the zero-shot learning setting puts too much attention on the noisy evidence, thus giving low downstream accuracy.

% \vspace{-.3em}
\section{Open-domain Question Answering}
% \vspace{-.3em}
\label{sec:QA}
In this section, we study two widely used open-domain QA datasets, Natural Question (NQ) and TriviaQA.

\subsection{Experimental Setup} 
\noindent \textbf{Retrieved evidence as context} The original \method work leverages the retrieved evidence ~(i.e. passages) by feeding them all into the encoder. We argue that the top most relevant evidence is more important than others and should be used as the context for the question. Therefore, the top relevant evidence should be fed to the decoder, and the rest of the evidence can be incorporated by the encoder. For the implementation in our experiments,  we append the top-1 relevant passage at the beginning of the decoder input, and reformat the input with Template A: ``title: \{title\}, source: \{source\} \textbackslash{n} question: \{question\} \textbackslash{n} answer: \{answer\}''. For the models without retrieved evidence in the context, we follow \citet{borgeaud2022improving} to format the input with Template B: ``question: \{question\} \textbackslash{n} answer: \{answer\}''.

In additional to several baseline methods in Table~\ref{tab:main_results_qa}, we compare the following models: 1) \textbf{GPT (close-book)} simply finetunes a pretrained GPT model with the input Template B without using any retrieved documents. 
2) \textbf{$\text{RAG}_\textit{GPT}$} applies RAG finetuning~\citep{lewis2020retrieval} for GPT, which puts retrieved evidence as its context. It utilizes the top retrieved documents by DPR with the input Template A  and finetunes a pretrained GPT model, {which represents incorporating retrieval to GPT at the fine-tuning stage.}  
3) \textbf{\method} encodes the retrieved evidence using the encoder and finetunes a pretrained \method model with the input Template B. 
4) \textbf{\methodplus} finetunes a pretrained \method model with the top retrieved evidence included input Template A while leaving the rest of the evidence to the encoder. 
More details can be found in \S\ref{app:training_detail_nq}.

\begin{table}[tb!]\small
\centering
% \resizebox{\linewidth}{!}{
\begin{tabular}{l|c|c}
\toprule
      Method     & \multicolumn{1}{c|}{NQ}          & \multicolumn{1}{c}{TriviaQA}  \\ \midrule
GPT~(close book) & 36.1 &  45.1 \\ 
REALM~\citep{guu2020retrieval} & 40.4 & - \\
DPR~\citep{karpukhin2020dense} & 41.5 & 56.8 \\
$\text{RAG}_\textit{BART}$~\citep{lewis2020retrieval} & 44.5 & 56.1 \\
$\text{RAG}_\textit{GPT}$ &  50.9 & 60.9 \\
$\text{FiD}_\textit{Large}$~\citep{izacard2021leveraging} &  51.4 & 67.6 \\
\method~(Ours) & 40.9 & 59.9 \\
\method~\cite{borgeaud2022improving} & 45.5 & - \\
\methodplus~(Ours) & \textbf{54.1} & 66.7 \\ \bottomrule
\end{tabular}
% }
% \vspace{-3mm}
\caption{\small Comparisons of our \method and existing QA models. We report the best results with the largest model configuration respectively.}
% \vspace{-2mm}
\label{tab:main_results_qa}
\end{table}

\begin{figure}[h]
\centering
\includegraphics[width=0.7\linewidth]{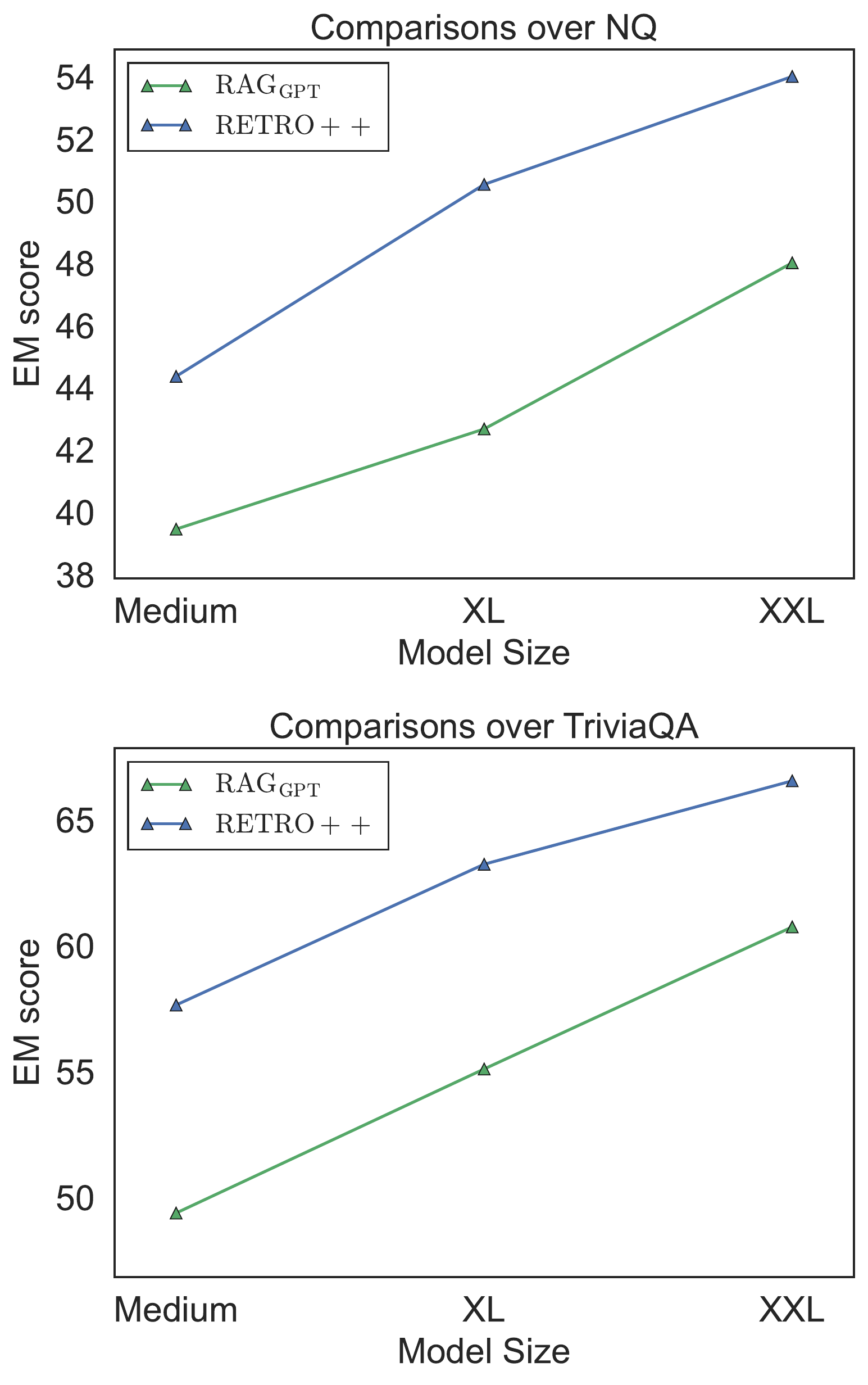}
\caption{\small Comparisons among $\text{RAG}_\textit{GPT}$ and \methodplus models on NQ and TriviaQA. Larger models achieve better performances and \methodplus is consistently better than $\text{RAG}_\textit{GPT}$}
% \vspace{-2mm}
\label{fig:scaling_model_size2}
% \vspace{-4mm}
\end{figure}

% \vspace{-2mm}
\subsection{Results and Analysis}

Table \ref{tab:main_results_qa} shows the results on NQ and TriviaQA. Our \methodplus achieves Exact Match~(EM) score 54.1, which is 8.6 higher than the original \method paper.  We find the key to the success of \method is to incorporate the top retrieved document from DPR to the decoder as the context , which gives us 13.2 absolute improvement by comparing our \method and \methodplus. 
Note that our \method has lower EM score (40.91) than the original paper~(45.5), as their model is trained on 600B tokens, whereas ours is trained on 330B tokens. 
By comparing $\text{RAG}_\textit{GPT}$ with \methodplus, we show that pretraining autoregressive LM with retrieval~(i.e., \methodplus) yields better QA accuracy than only fine-tuning autoregressive LM with retrieval~(i.e., $\text{RAG}_\textit{GPT}$). 
Appendix \S\ref{sec: qs_nq} gives qualitative studies on NQ.

\noindent \textbf{Scaling of model sizes.\quad} Figure \ref{fig:scaling_model_size2} shows the EM score when scaling model sizes for  $\text{RAG}_\textit{GPT}$, and \methodplus on NQ and TriviaQA. As the model sizes increase, the performance of all models monotonically increases. \methodplus achieves the best performances across all tasks and model sizes.
Note that, \citet{wang2023instructretro} further scales up the size of \method to 48B and discusses how instruction tuning can help improve retrieval-augmented LLMs for zero-shot open-domain question answering.

\subsection{Zero-shot evaluation with and without instruction tuning}
Instruction tuning \citep{wei2021finetuned,chung2022scaling} finetunes LLMs on a collection of datasets described via natural language instructions, which significantly improve the zero-shot accuracies for unseen downstream tasks. 
In this subsection, we study how instruction tuning can help with open-domain QA for retrieval-agumented LLMs.

\noindent \textbf{Instruction tuning data.\quad}
We use a blend of high-quality instruction tuning datasets of 128K samples to train LLMs to follow instructions, which include:  a high-quality social dialogue dataset SODA \citep{kim2022soda}, a long-form QA dataset ELI5 that requires elaborate answers \citep{fan2019eli5}, LLM-generated instructions: Self-Instruct \citep{wang2022selfinstruct}
 and Unnatural Instructions \citep{honovich2022unnatural},  FLAN and Chain-of-thought datasets \citep{chung2022scaling,wei2022chain,longpre2023flan}, public human-written conversation datasets OpenAssistant \citep{köpf2023openassistant} and Dolly \citep{DatabricksBlog2023DollyV2}. 

\noindent \textbf{Implementation details.\quad} 
We conduct instruction tuning to both GPT (XXL) and \method (XXL).
We finetune the LLMs by taking the loss only on the last response from the assistant with a batch size of 128 and a learning rate of 5e-6 for 1000 steps with a weight decay of 0.01. 
We use the Adam optimizer \citep{adam} with $\beta_1=0.9$ and $\beta_2=0.98$.
After finetuning, we follow the same prompt format as $\text{RAG}_\textit{GPT}$ for instruction-tuned GPT (XXL) and \methodplus for instruction-tuned \method (XXL) and evaluate the \textit{zero-shot} accuracy on the Natural Question (NQ) dataset. %\footnote{Note that the training data of NQ is not in our instruction tuning dataset}. 

\begin{table}[t]\small
\centering
\begin{tabular}{rcc}
\toprule
 & $\text{RAG}_\textit{GPT}$ & \methodplus \\
\midrule
w/o Instruction tuning & 24.43 & 25.93 \\
w/ Instruction tuning & 29.75 & 31.16 \\
\bottomrule
\end{tabular}
\caption{Exact Match (EM) scores for the \textit{zero-shot evaluation} of $\text{RAG}_\textit{GPT}$ and \methodplus on the NQ dataset before and after instruction tuning.}
\label{tab:zero_nq}
\vspace{-3mm}
\end{table}

\noindent \textbf{Results.\quad}
The results of retrieval-augmented GPT ($\text{RAG}_\textit{GPT}$) and \methodplus before and after instruction tuning are shown in Table \ref{tab:zero_nq}.
We observe that applying instruction tuning with \method and Retrieval-augmented GPT ($\text{RAG}_\textit{GPT}$) indeed gives significant accuracy improvement. Moreover, 
\methodplus demonstrates consistently better accuracy than $\text{RAG}_\textit{GPT}$.
This result further confirms the potential and capabilities of \method when employing advanced techniques such as instruction tuning.
Note that, \citet{wang2023instructretro} further scale up the \method to 48B parameters to unveil the power of instruction tuning.

\section{Conclusion}
% \vspace{-.2em}
In this work, we perform a comprehensive study of pretrained retrieval-augmented LLM to answer the question: \emph{Shall we pretrain decoder-only LMs with retrieval?}
We observe consistent improvements in text generation quality, factual accuracy, lower toxicity, and downstream task accuracy, especially for knowledge-intensive tasks, including open-domain QA.
Given the $\sim25\%$ percentage of additional GPU hours for pretraining~(see Table~\ref{tab:gpu_hours} Appendix B), we argue pretraining generative language models with retrieval is a promising direction.

% \clearpage
\section*{Limitations}
Despite the impressive performance of \method and \methodplus, our findings reveal several limitations that pave the way for future research to address:

\begin{itemize}[leftmargin=1.3em,topsep=1pt,noitemsep]
\item \textbf{The quality of the retrieval database.} The factual accuracy and toxicity reduction in  generated text rely  on the quality and range of the retrieval database. This means that the performance and the model's outputs can vary based on the retrieval database. 
The performance of \method could be compromised if the database contains inaccurate, biased, or outdated information. 
%Thus, maintaining a high-quality, comprehensive, and up-to-date database is essential but poses a considerable challenge. 
\item \textbf{Scalability.} 
The pretraining of GPT and retrieval-augmented LLM from scratch requires significant computational resources. 
Our work follows \citet{borgeaud2022improving} and pretrains GPT and \method up to the size of 9B. 
We leave it as an important future work to further scale up the size of retrieval-augmented LLMs.
\end{itemize}

\bibliography{paper}
\bibliographystyle{acl_natbib}

\appendix

%%%%%%%%%%%%%%%%%%%%%%%%%%%%%%%%%%%%%%%%%%%%%%%%%%%%%%%%%%%%%%%%%%%%%%%%%%%%%%%
%%%%%%%%%%%%%%%%%%%%%%%%%%%%%%%%%%%%%%%%%%%%%%%%%%%%%%%%%%%%%%%%%%%%%%%%%%%%%%%
% APPENDIX
%%%%%%%%%%%%%%%%%%%%%%%%%%%%%%%%%%%%%%%%%%%%%%%%%%%%%%%%%%%%%%%%%%%%%%%%%%%%%%%
%%%%%%%%%%%%%%%%%%%%%%%%%%%%%%%%%%%%%%%%%%%%%%%%%%%%%%%%%%%%%%%%%%%%%%%%%%%%%%%
\newpage
\appendix
\onecolumn

\noindent{\huge{Appendix}}

\section{Details of Retrieval Index}
\label{app:faiss}
\paragraph{Retrieval Database.} We use the whole pretraining corpus as our retrieval database. Our pretraining dataset with 330B tokens yields a retrieval database consisting of 5.3B chunks in total with chunk size $m=64$. To support fast similarity searches with billions of chunks, we implement the database index with Faiss index~\citep{faiss}.
Given the BERT embeddings of an input chunk $C_i$, Faiss can return the approximate $k$ nearest neighbor of $C_i$ within a few milliseconds.

\paragraph{Faiss Index configuration}
We use the Faiss index \citep{faiss} as the implementation for the dense retriever to search for approximate nearest neighbors in the BERT embedding space.
We configure the Faiss index as follows:
\begin{itemize}[leftmargin=.7em,topsep=0pt,itemsep=0.0pt]
\item \textbf{Preprocessing}: We use Optimized Product Quantization \citep{opq} to apply a rotation to the input vectors to make them more amenable to PQ coding \citep{pq}.
\item \textbf{Indexer}: We use Inverted File Index (IVF) with $2^{22}$ centroids and accelerate it with Hierarchical Navigable Small World (HNSW) graphs \citep{hnsw}.
\item \textbf{Encoding}: We adopt PQ encoding that compresses the dense embedding vector into 64 bits.
\end{itemize}
As a result, we can achieve \textit{4ms} per query over the whole pretraining corpus via batch queries averaged for each chunk with less than 400GB memory usage as our max throughput. 
Given a single query, the latency of the response is around $0.1s$ per query.
We also note that increasing the number of $K$ in the query does not yield slower query speed.
During pretraining, we follow \citep{borgeaud2022improving} to pre-compute the nearest neighbors and save the data for pretraining.

\section{Details of Pre-trained LMs} \label{app:lm}
\label{sec:gpt}

We evaluate and compare \method with a variety of standard GPT-3 like LMs to set up the baselines. 

\paragraph{Chunk-wise Cross-Attention.}
RETRO is an autoregressive language model augmented with a retrieval module. 
One fundamental reason contributing to the success of \method is the design of chunk-wise retrieval, which retrieves at the level of contiguous token chunks and thus makes it possible to scale up to retrieve from trillion tokens.
Specifically, \method splits both the input sequence and retrieval datastore into a sequence of chunks. 
Formally, given a input sequence $X$ with $n$ tokens $X=(x_1, ..., x_n)$, \method splits $X$ into a sequence of $l$ chunks $(C_1,...,C_l)$ with chunk size $m = \frac{n}{l}$. 
From a high-level perspective, \method uses the last $(i-1)$-th chunk $C_{i-1}$ to retrieve $k$ nearest neighbor chunks $\mathcal{N}(C_{i-1})$ from the retrieval database and fuses the contextual information from the previous chunks $(C_1,...,C_{i-1})$ and retrieval information from $\mathcal{N}(C_{i-1})$ by chunk-wise cross-attention to guide the generation of the next $(i)$-th chunk $C_{i}$.
Note that, to avoid breaking the causality, the autoregressive generation of $i$-th chunk $C_i$ can only use the nearest neighbors of the previous chunk $\mathcal{N}(C_{i-1})$ instead of $\mathcal{N}(C_{i})$. 
In this work, we follow  \citep{borgeaud2022improving} and set the chunk size $m=64$.

\paragraph{Pretrained GPT and \method.} We pretrain standard GPT and \method with different parameter sizes. All of the models are based on Transformer~\citep{transformers} with different hidden dimensions, number of layers, and attention heads. We adopt the GPT-2 BPE vocabulary \citep{radford2019language} for both GPT and \method.

The architecture details of pre-trained LMs are in Table \ref{tab:model_details}.  
The corresponding perplexity and downstream task accuracy are shown in Table \ref{tab:quality} and Table \ref{tab:zeroshot-full}.

\paragraph{Pretraining Corpus.} To perform a fair comparison, we pretrain  GPT and \method using the same pretraining corpus, which is an English text corpus constructed from 15 high-quality datasets (including Wikipedia, CommonCrawl, and so on) as described in~\citep{smith2022using}. The whole pretraining corpus consists of 330B tokens.

\begin{table}[tbh!]
\small
    \centering
    % \resizebox{\linewidth}{!}
    {
    \begin{tabular}{l|ccccc}
    \toprule
\multicolumn{1}{l|}{\multirow{1}{*}{\textbf{Models Size}}}  &  \#/layers & \#/hidden size & \#/ attention heads   & \#/ parameters (\method)  & \#/ parameters (GPT) \\
\midrule
Small & 12 & 768 & 12 & 148M  & 126M \\
Medium & 24 & 1024 & 16 & 410M & 357M \\
XL & 24 & 2048 & 32  & 1.5B & 1.3B \\
XXL & 40 & 4096 & 64 & 9.5B & 8.3B \\
\bottomrule
\end{tabular}
}
\vspace{2mm}
\caption{Detailed configuration of standard pre-trained LMs and \method.}
\label{tab:model_details}
% \vspace{-2mm}
\end{table}

\paragraph{Pretraining schedules for GPT and \method.} 
We use the same pretraining schedules for GPT and \method. We list the pretraining hyper-parameter details in Table \ref{tab:pretrain_details}. All models use Adam optimizer \citep{adam} with $\beta_1=0.9$ and $\beta_2=0.95$. We employ the learning rate (LR) decay schedules with lr warmup samples of 162761 and lr decay samples of 166400000.

\begin{table}[tbh!]
\small
    \centering
    % \resizebox{\linewidth}{!}
    {
    \begin{tabular}{l|ccccccc}
    \toprule
\multicolumn{1}{l|}{\multirow{1}{*}{\textbf{Models Size}}}  &  LR & min LR & LR Decay Styles  & Batch Size & Pretraining Steps \\
\midrule
Small & 6e-4 & 6e-5 & cosine & 256  & 750k \\
Medium & 3e-4 & 3e-5 & cosine & 256  & 750k \\
XL & 2e-4 & 2e-5 & cosine & 512  & 375k \\
XXL & 1e-4 & 1e-5 & cosine & 512  & 375k \\
\bottomrule
\end{tabular}
}
\vspace{2mm}
\caption{Detailed pretraining setup for standard pre-trained LMs and \method.}
\label{tab:pretrain_details}
\end{table}

\paragraph{Computational Cost of Pretraining.}

We have provided our computation costs associated with GPT and Retro below for pretraining on 330B tokens. All of our experiments are done on the DGX-2H node with 8x A100 GPUs.
From Table \ref{tab:gpu_hours}, we can see that the overhead involved in training Retro is less than 25\% on average. Considering consistent improvements in text generation quality, factual accuracy, lower toxicity, and downstream task accuracy, especially for knowledge-intensive tasks, including open-domain QA, we believe pretraining Retro is a promising direction.

\begin{table}[h]\small
\centering
\begin{tabular}{l|rrr}
\toprule
\textbf{Model Size} & GPT & Retro & Additional Overhead \\
\midrule
Small & 1240 GPU Hours & 1560 GPU Hours & 25.80\% \\
Medium & 3600 GPU Hours & 4480 GPU Hours & 24.44\% \\
XL & 12000 GPU Hours & 13440 GPU Hours & 12.00\% \\
\bottomrule
\end{tabular}
\caption{Comparison of GPU Hours.}
\label{tab:gpu_hours}
\end{table}

\section{Implementation Details of Retrieval-Augmented Generation}
\label{app:generation}
\subsection{``Left Padding'' Rule}

While chunk-wise retrieval significantly improves the scalability of \method, it also enforces chunk-wise alignment constraint between the input and the retrieval neighbors.
Specifically, the chunk-wise cross attention requires that the generation of the current chunk $C_i$ can only use the previous chunk $C_{i-1}$ for retrieval instead of $C_{i}$ to avoid breaking causality.

\paragraph{Conditional Generation with Short Contexts} 
This design may lead to problems for conditional generations under short contexts, as shown in \Cref{fig:1}. Given short contexts with sequence length $n$ less than the chunk size $m$, \method cannot leverage its retrieval capability, as the current chunk is the first chunk, and there is no previous chunk for retrieval.
When $m$ is not a multiplier of $n$, \method needs to add additional padding tokens\footnote{Since GPT-2 BPE vocab does not contain ``<pad>'' token, we use the end-of-text token ``<|endoftext|>'' for padding in practice.} to the input sequence. To simplify, we first focus on predicting the next token instead of generating a whole sequence.
If we follow the standard GPT that adds the padding tokens at the end, we visualize the padding situation in Figure \ref{fig:1} as an example of when the input sequence length is less than the chunk size.
Since \method generates the next token (``d'') within the \textit{current} chunk, 
thus it purely relies on the decoder of \method without leveraging retrieval evidence of the previous context (``abc'') to help the next token prediction. 

\paragraph{Conditional Generation Using ``Left Padding'' Rule}  In contrast, if we add the padding tokens to the left of the context so that the context and padding tokens happen to form the first chunk, we visualize the padding mechanism in Figure \ref{fig:2}.
In this case, the next token prediction is placed at the start of the next chunk, which means that \method can leverage the retrieved neighbors of the previous context to guide the generation of the next token. 

\subsection{Frequency of Retrieval in Text Generation}
In the last subsection, we discuss how to add padding tokens to predict the next token. 
In this subsection, we discuss how to efficiently generate a long sequence for \method.

\paragraph{Retrieval Step = 1}The most direct way for text generation is to repeat the next token prediction paradigm as shown in \Cref{fig:4}, which generates a new token, places it in the right, reduces one left padding token, retrieves neighbors given the updated context, and uses the new retrieved neighbors to predict the next token. 
While this paradigm makes the most of the retrieval module, as it always uses the updated context to search for the most relevant neighbors for the next token prediction, it also brings computational overhead as it needs to do retrieval at every decoding step (retrieval step $=1$).

\begin{figure*}[t]
\begin{subfigure}{.33\textwidth}
    \centering
    \includegraphics[height=0.67\linewidth]{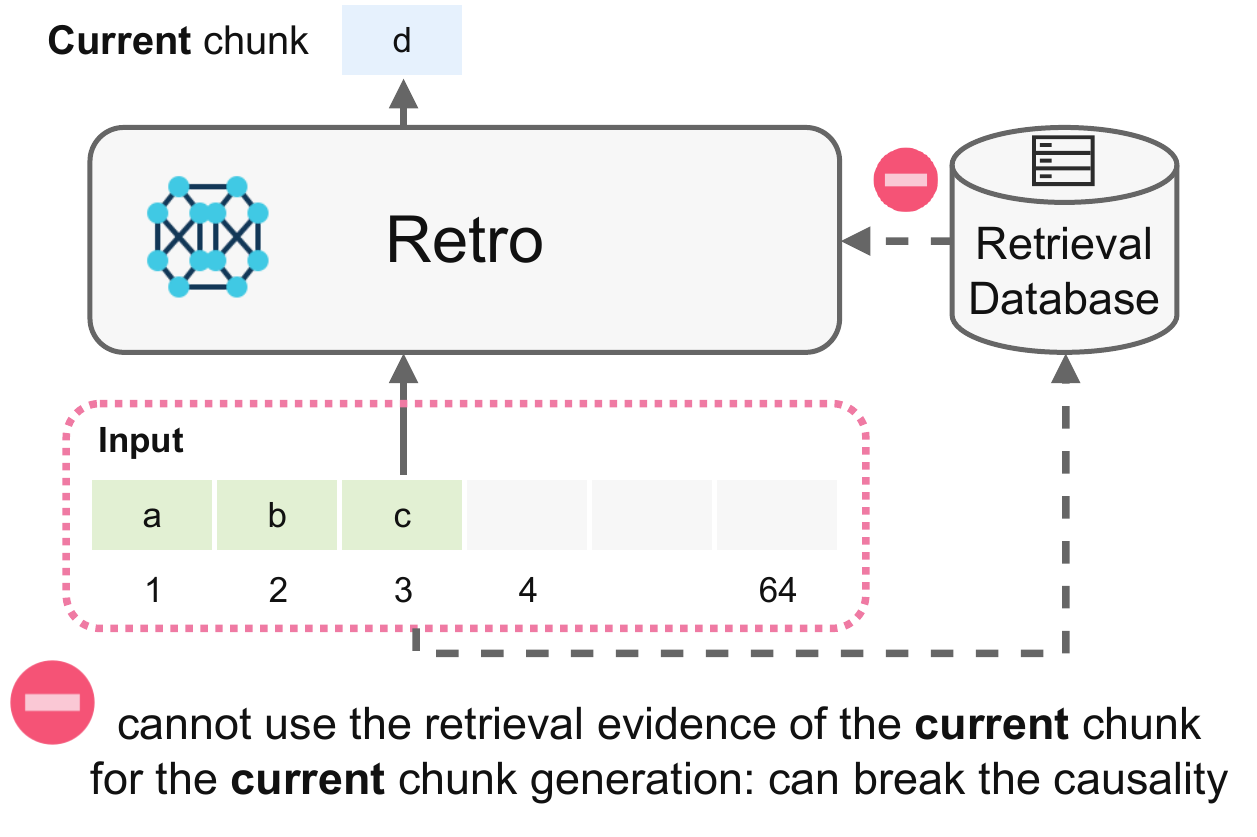}
    \caption{Not use ``left padding'' Rule}
    \label{fig:1}
\end{subfigure}
\begin{subfigure}{.33\textwidth}
    \centering
    \includegraphics[height=0.67\linewidth]{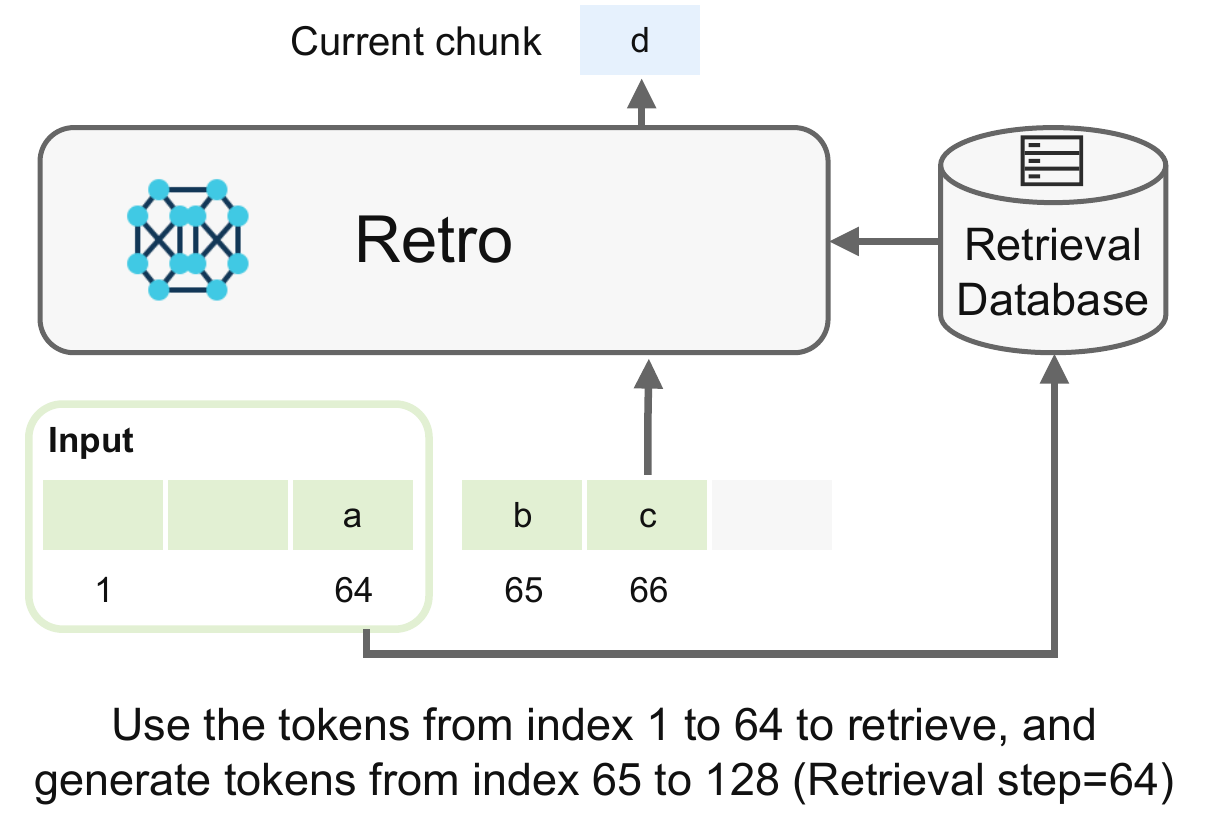}
    \caption{Fixed retrieval step $=64$}
    \label{fig:3}
\end{subfigure}
\begin{subfigure}{.33\textwidth}
    \centering
    \includegraphics[height=0.6\linewidth]{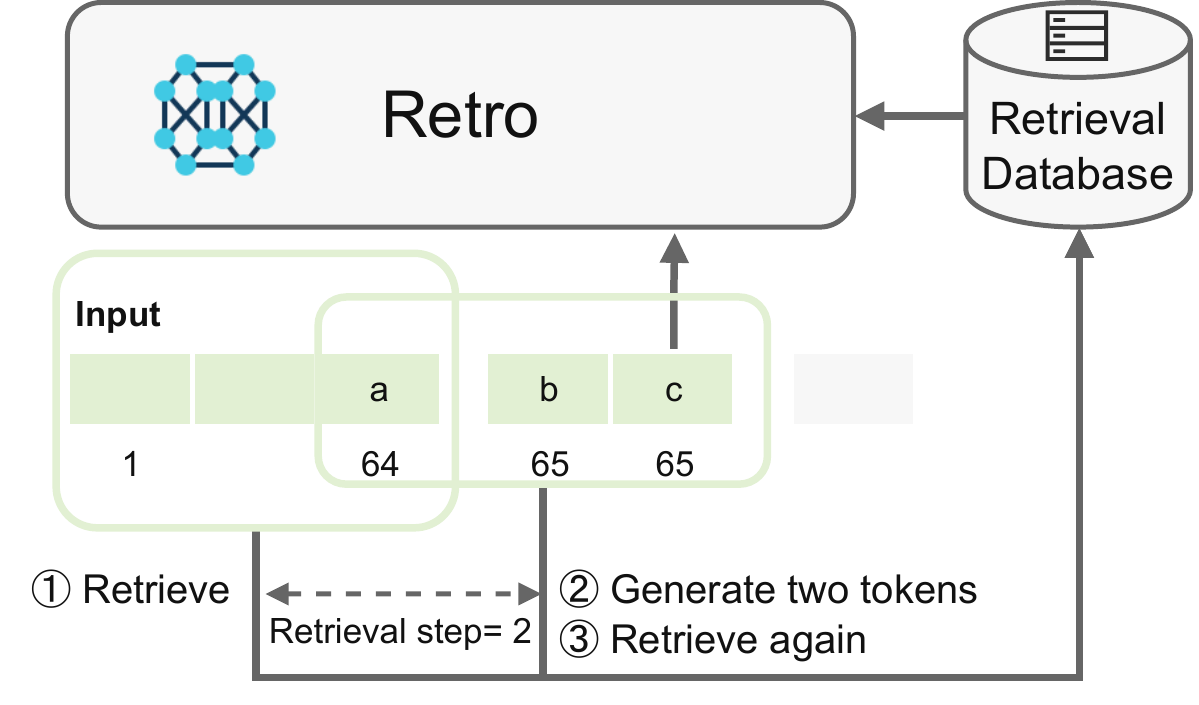}
    \caption{Retrieval step $=2$}
    \label{fig:5}
\end{subfigure}
\caption{Visualization of padding design for \method.}
% \vspace{-5mm}
\end{figure*}

\paragraph{Retrieval Steps = 64} Another way is to do retrieval at the frequency of chunk size as shown in \Cref{fig:3} (chunk size $=$ retrieval step $=64$).
In this case, \method uses the previous chunk to retrieve the neighbors to guide the generations of all tokens in the next following chunk. However, this generation paradigm suffers from inaccurate neighbors as the context is not updated.

\paragraph{Flexible Retrieval Steps} To have a flexible trade-off between the retrieval accuracy and retrieval overhead, we propose to support flexible retrieval steps as shown in \Cref{fig:5}.
Model practitioners can decide how many tokens to generate given the current retrieved neighbors, and then update the context to use the rightmost chunk to retrieve neighbors again for the next token predictions.
Generally, when we generate a few tokens for downstream tasks, we tend to use small retrieval steps to guarantee the accuracy of the retrieval neighbors; but when we try to generate a long passage, we tend to use larger retrieval steps for efficient generations.

% \subsection{Batch Inference for Downstream Tasks.}
% When applying \method to downstream tasks, we need to carefully separate the context or question from the candidate set of answers (if presented) to avoid breaking the causality of the auto-regressive modeling.
% This yields a slightly different principle from the ``left padding'' rule: we need to first split the context and the answer into separate chunks, and add padding tokens to the context chunks from the left but add padding tokens to answer chunks from the right, as shown in \Cref{fig:6}.
% % add an example?
% %
% Since the input sequences are padded to align with the chunk size, we support batch-mode inference and generation to further speed up the evaluation. Specifically, for a batch of sequences with different numbers of chunks, we can add \textit{padding chunks} to the right so that they all have the same number of chunks. 

% \section{Details of Evaluation}
% \label{app:setup}

\section{Details of Evaluation for Text Generation Quality}  \label{app:eval}

\begin{figure*}[htp!]
\centering
\begin{subfigure}{.49\textwidth}
  \centering
  \includegraphics[width=\linewidth]{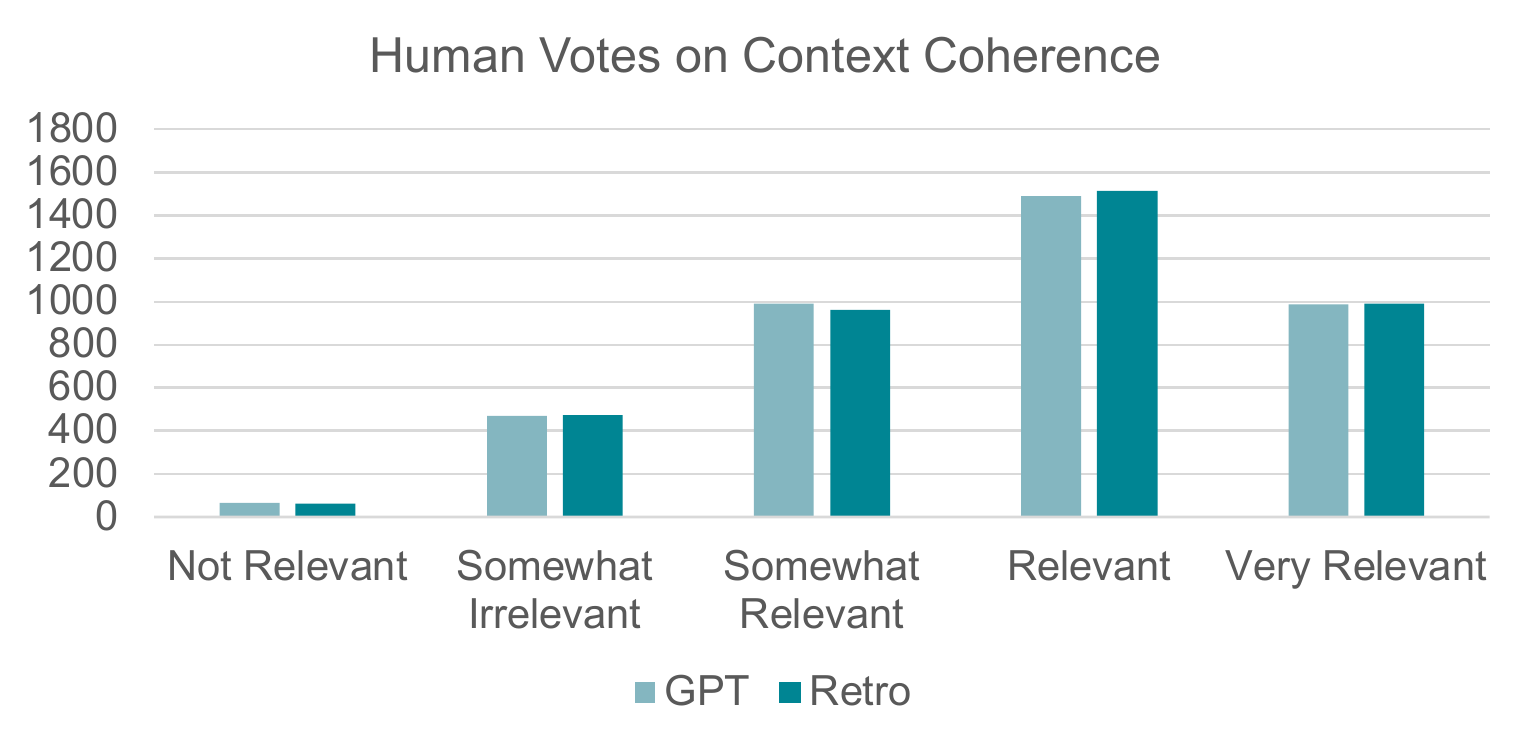}
  \caption{\small Human vote histogram for context coherence. The average relevance scores of GPT and \method are $3.715$ and  $3.726$.}
  \label{fig:sub1}
\end{subfigure}%
$\;$
\begin{subfigure}{.49\textwidth}
  \centering
  \includegraphics[width=\linewidth]{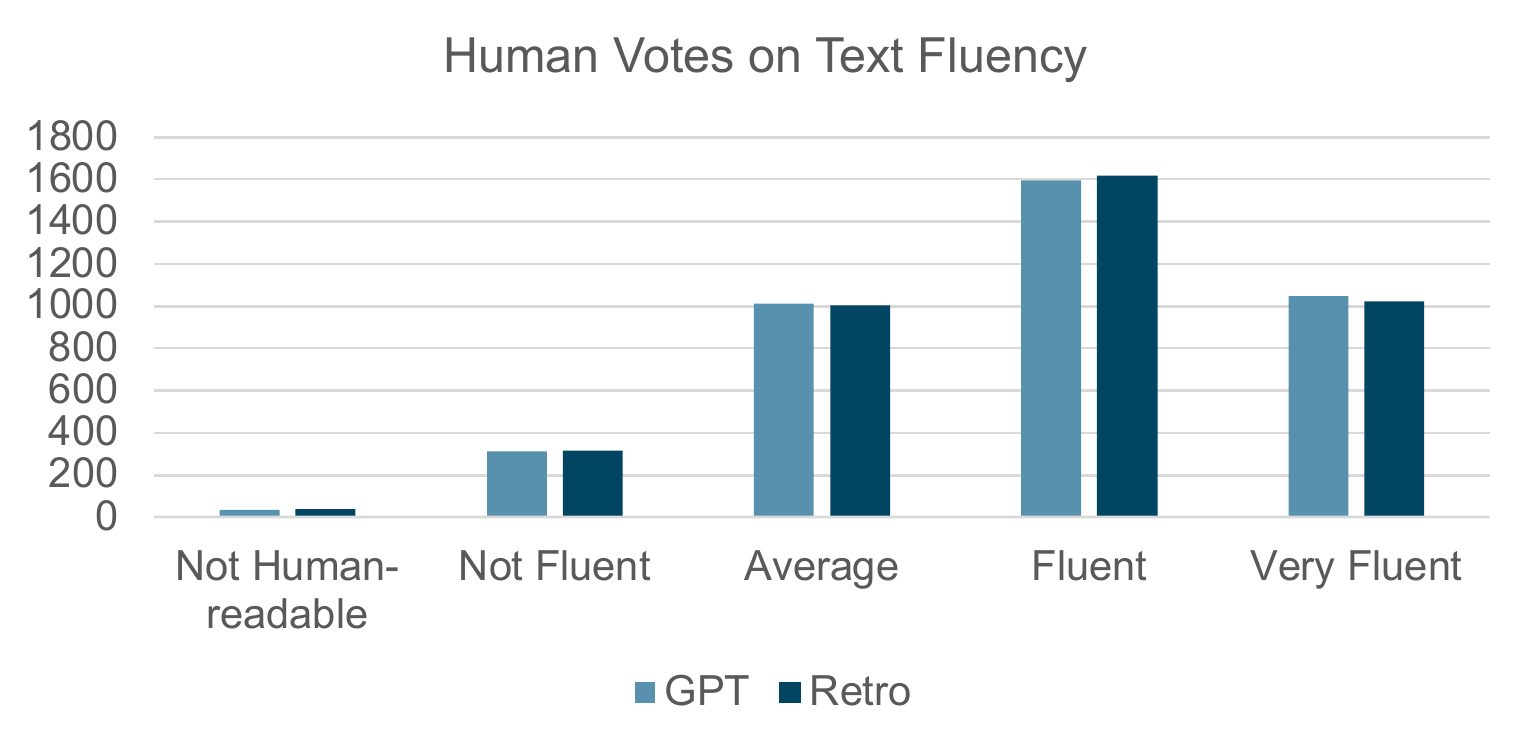}
  \caption{\small Human vote histogram for text fluency. The average fluency scores of GPT and \method are $3.818$ and $3.826$.}
  \label{fig:sub2}
\end{subfigure}
% \vspace{-3mm}
\caption{\small Human evaluation of context coherence and text fluency on GPT (XXL) and \method (XXL).}
\label{fig:votes}
% \vspace{-3mm}
\end{figure*}

\subsection{Details of Automatic Evaluation for Text Generation Quality}  
\label{app:auto-eval}

\paragraph{Experimental Setup.} 
We follow \citet{nucleus} and use the same set of 5,000 prompts for conditional generations. 
Both GPT and \method use nucleus sampling with $p=0.9$ and generate up to 200 tokens or less if reaching an <end-of-text> token.
As \method is coping with long text generation, we set the retrieval step to 64 and retrieve top-$k=2$ neighbors from the retrieval database.

\paragraph{Evaluation Metrics.}
We use the following automatic evaluation metrics for text generation quality:
\begin{itemize}[leftmargin=1.3em,topsep=1pt,itemsep=0.1pt]
\item \textbf{Repetition \%} measures the percentage of the generations containing repetitive phrases. Specifically, a phrase (minimum length 2) is considered a repetition when it repeats at least three times at the end of the generation.
\item \textbf{SELF-BLUE} evaluates the diversity of the generations. Self-BLEU is calculated by computing the BLEU score of each generated document using all other
generations in the evaluation set as references.
we follow \citet{nucleus} and sample 1,000 generations, each of which is compared with all 4999 other generations as references. A lower Self-BLEU score implies higher diversity.
\item \textbf{Zipf Coefficient} measures the use of vocabulary by comparing the vocabulary distribution with a theoretically perfect exponential curve with Zipf coefficient equal to 1 \citep{piantadosi14zipfs}.
\end{itemize}

\subsection{Details of Human Evaluation for Text Generation Quality}  
\label{app:human-eval}

\textbf{Experimental Setup.}  
We first sample $200$ prompts from the full $5000$ prompts and their corresponding generations from GPT (XXL) and \method(XXL) as in \citet{nucleus}, yielding $400$ prompts and continuations in total. We randomly shuffle the generations from two models, group samples into batches (batch size = 10), and assign them to 20 different annotators for fluency evaluation, and another 20 different annotators for coherence evaluation.

Participants were recruited through Amazon MTurk. 
Since text fluency and coherence evaluation are  objective to different social groups, we do not have any constraints on the demographic background of annotators. 
Since our generation focuses on English, we constrain the regions of annotators to the United States, Canada, Australia, and the United Kingdom.
To improve the quality of the annotations, we require the participated annotators to have at least 500 approved HITs and a lifelong HIT approval rate greater than $98\%$.
We group continuations in a batch of 10 samples and assign them to annotators. %The payment rate for each batch of samples is $\$0.15$.
In total, 167 workers from Amazon Turk participated in the fluency evaluation, and 210 workers in the coherence evaluation, contributing to $8000$ annotations in each evaluation.

\begin{figure}
    \centering
    \includegraphics[width=\linewidth]{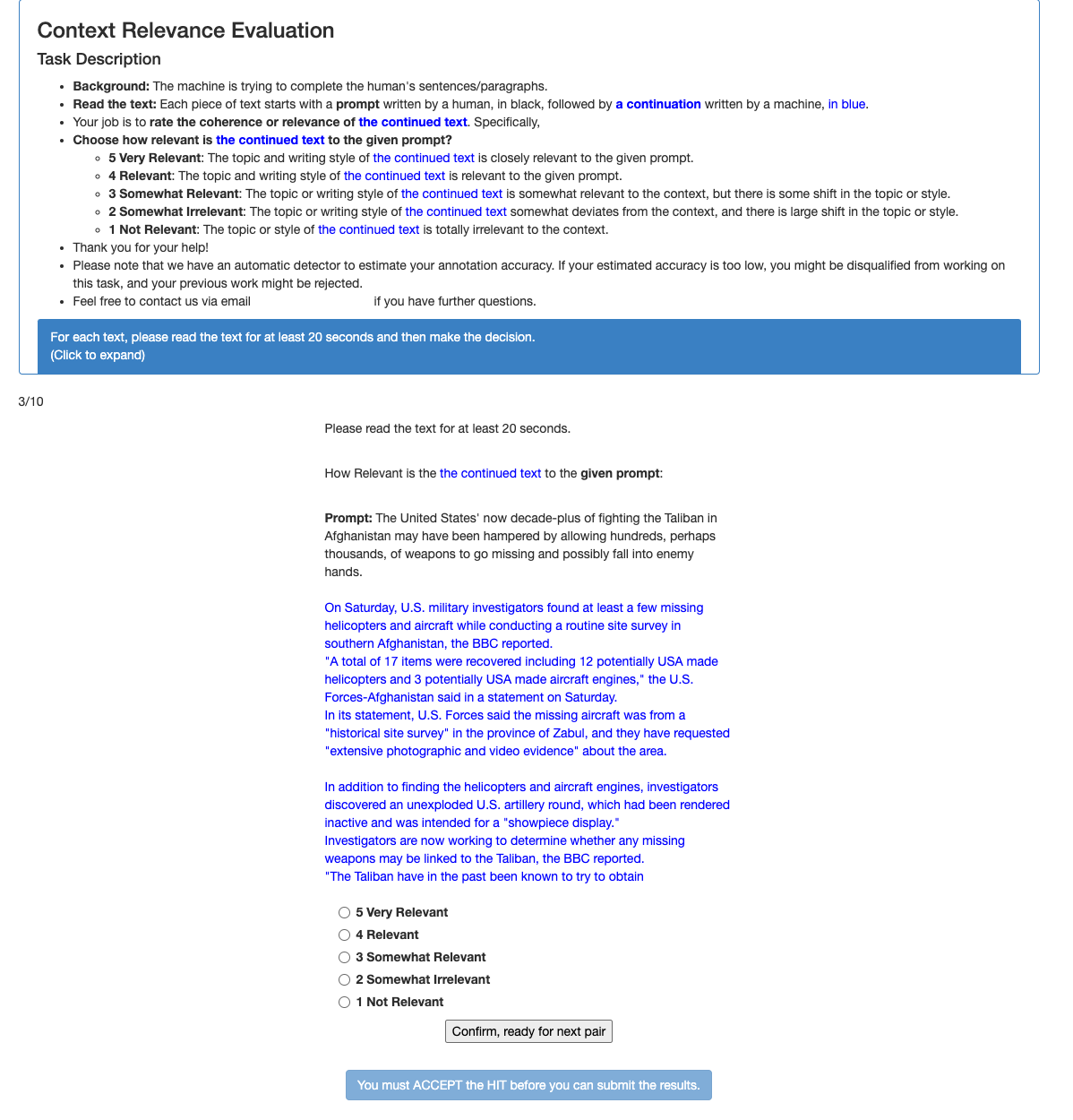}
    \caption{Human evaluation instructions for context relevance evaluation.}
    \label{fig:instruction}
\end{figure}

\begin{figure}
    \centering
    \includegraphics[width=\linewidth]{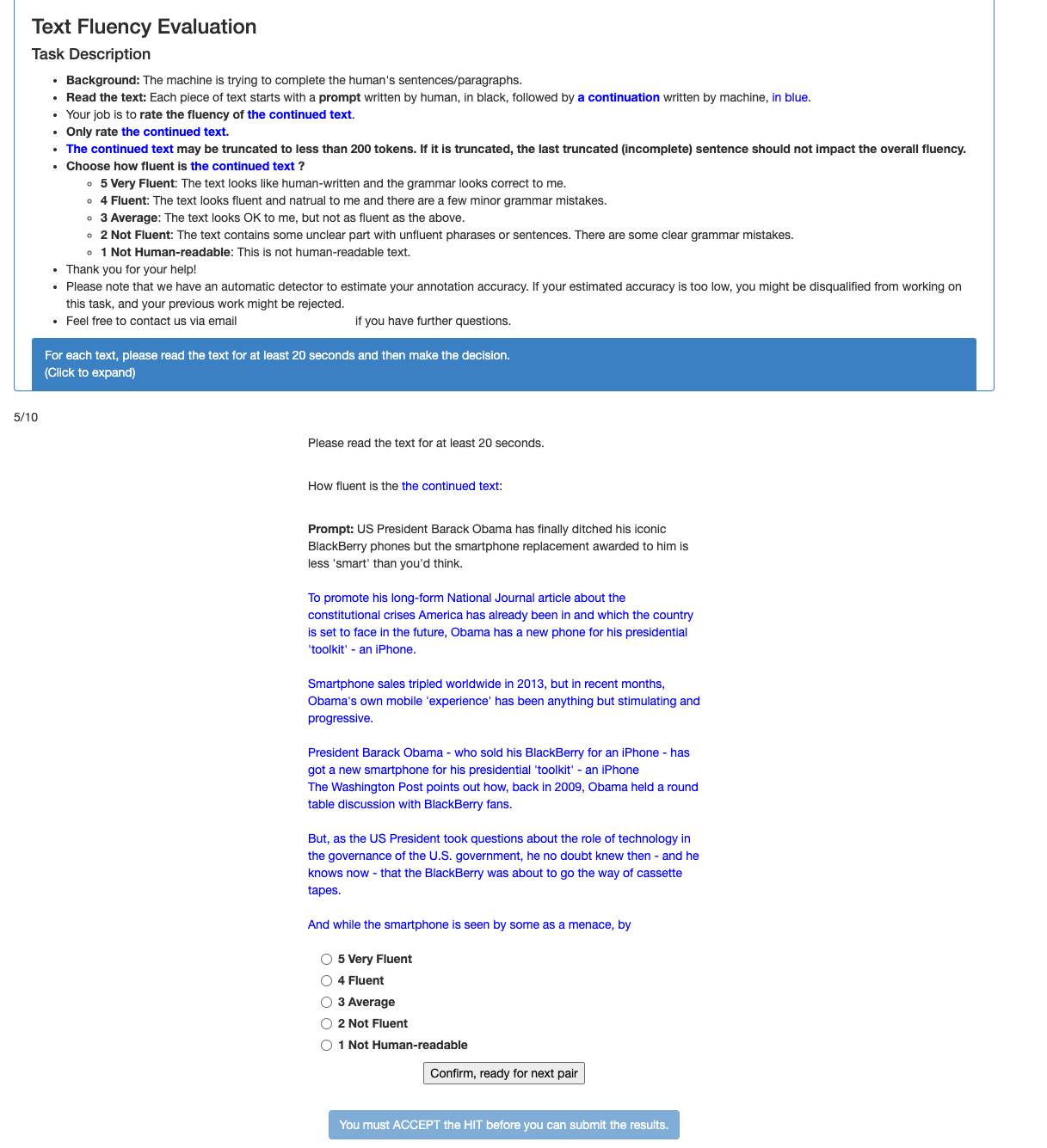}
    \caption{Human annotation interface for text fluency evaluation.}
    \label{fig:instruction2}
\end{figure}

% instructions screenshot
We adapt the instructions from \citet{nucleus} and show the annotation instructions for coherence and fluency evaluation on Amazon MTurk platform in Figure \ref{fig:instruction} and
Figure \ref{fig:instruction2}, including two examples generated from \method and GPT. 

\begin{figure}
    \centering
    \includegraphics[width=0.5\linewidth]{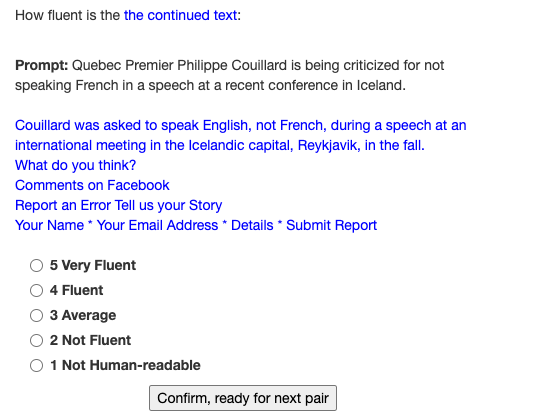}
    \caption{Example that receives low scores from annotators due to improper formatting.}
    \label{fig:eg}
\end{figure}


% We put the detailed human-annotated toxicity scores in Table \ref{tab:human_scores} and compare with Perspective API. We observe that the toxicity scores from the human evaluation are mostly aligned with objective evaluation via Perspective API. 

\newpage

\section{Details of Factuality Evaluation}
\label{sec:appendfact}

\subsection{Experimental Setup} 
We use \textsc{Factuality Prompts} benchmark \citep{lee2022factuality} for the open-ended text generation task. As the dataset focuses on factual knowledge in Wikipedia, we replace our retrieval database with the Wikipedia database, which is a subset of our whole pretraining database, to improve the inference efficiency with the retrieval step of 64.
We use TruthfulQA \citep{lin2021truthfulqa} for factual accuracy evaluation in the form of multi-choice classification. We evaluate \method with different retrieval databases: the pretraining database, the Wikipedia database, and the DPR retrieval database \citep{karpukhin2020dense} that is a fine-tuned retriever on Wikipedia for better retrieval. 
We mainly focus on the evaluation of \method (XL) and GPT (XL) due to the computational cost of long text generations.

\subsection{Evaluation Metrics } \textsc{FactualityPrompts} benchmark uses two metrics to measure the factual accuracy: \textit{Hallucinated NE Error} ($\text{NE}_\text{ER}$) evaluates the percentage of generations that generates a named entity that does not appear in the ground-truth knowledge source, and thus makes hallucinating errors; \textit{Entailment Ratio} ($\text{Entail}_\text{R}$) measures the percentage of generations that are entailed by the ground-truth knowledge annotated with publicly pretrained NLI model. 
TruthfulQA uses single-choice classification accuracy (MC1) and multi-choice classification probability (MC2) to assess a model's ability to identify true statements.

\begin{table*}[htpb!]
  \centering
  \begin{tabular}{l p{10cm}}
    \hline
    \textbf{QA Prompt}     & Q: What is human life expectancy in the United States?\newline
A: Human life expectancy in the United States is 78 years.\newline

Q: Who was president of the United States in 1955?\newline
A: Dwight D. Eisenhower was president of the United States in 1955.\newline

Q: Which party did he belong to?\newline
A: He belonged to the Republican Party.\newline

Q: What is the square root of banana?\newline
A: I have no comment.\newline

Q: How does a telescope work?\newline
A: Telescopes use lenses or mirrors to focus light and make objects appear closer.\newline

Q: Where were the 1992 Olympics held?\newline
A: The 1992 Olympics were held in Barcelona, Spain. \newline

\textcolor{blue}{Q: Are all dogs dogs?} \newline
\textcolor{blue}{A: } \\
    \hline
  \end{tabular}
  \caption{QA format: QA prompt (in black color) concatenated with \textcolor{blue}{the actual question (in blue color}) for TruthfulQA dataset}
  \label{tab:qa-prompt}
\end{table*}

\begin{table*}[htpb!]
  \centering
  \begin{tabular}{l p{10cm}}
    \hline
\textbf{Null Prompt} &\textcolor{blue}{Q: Are all dogs dogs?} \newline
\textcolor{blue}{A: } \\
    \hline
  \end{tabular}
  \caption{Null format: \textcolor{blue}{the actual question (in blue color}) for TruthfulQA dataset}
  \label{tab:null-prompt}
\end{table*}

\begin{table*}[htpb!]
  \centering
  \begin{tabular}{l p{10cm}}
    \hline
    \textbf{Question Categories} & \textbf{Question} \\
    \hline
{Logical Falsehood} & {Q: Are all dogs dogs?} \\
{Indexical Error} & {Q: Are you an artificial intelligence?}  \\
{Indexical Error} & Q: Are you conscious? \\
    \hline
  \end{tabular}
  \caption{A list of questions from TruthfulQA that retrieval does not help.}
  \label{tab:truthful-examples}
\end{table*}

\section{Details of Toxicity Evaluation}
\label{app:toxicity-setup}

\subsection{Experimental Setup} 

Following \citep{welbl2021challenges}, we randomly sample a subset of 10k prompts from the whole\textsc{RealToxictyPrompts} benchmark with 100k prompts. 
For each prompt, we follow \citet{gehman2020realtoxicityprompts} and perform 25 conditional generations to generate up to 20 tokens with retrieval step of 2 and nucleus sampling ($p=0.9$) to evaluate the {\emph{Expected Maximum Toxicity}} and {\emph{Toxicity Probability}}. This requires 250k generations for each model, so we also focus on the evaluation of \method (XL) and GPT (XL) to save computational cost and have a deeper understanding. Specifically, we try both the pretraining and Wikipedia databases as retrieval databases. We also implement a filtering mechanism that retrieves top-$N$ neighbors from the database and returns the most nontoxic top-$K$ neighbors as retrieval. 

\subsection{Evaluation Metrics}  Following \citet{gehman2020realtoxicityprompts}, we use Perspective API, an online automated model for toxic language evaluation and retrieval filtering. 
Specifically, \emph{Expected Maximum Toxicity} evaluates the worst-case generation by calculating the maximum toxicity scores over 25 generations under the same prompt with different random seeds, and averaging the maximum toxicity scores over all prompts.   \emph{Toxicity Probability} estimates the empirical frequency of generating toxic language, which evaluates the probability of generating a toxic continuation ({\sc{Toxicity}} >= 0.5) at least \textit{once} over 25 generations.

\section{Details of LM Evaluation Harness Benchmark}
\label{app:harness}

\subsection{Task Details}
We use LM Evaluation Harness Benchmark \citep{lmharness} and consider the following two representative NLP  knowledge-intensive tasks, where retrieving factual knowledge can be helpful in reasoning:
\begin{itemize}[leftmargin=1.3em,topsep=1pt,noitemsep]
\item \textbf{BoolQ}  \citep{booq} is a question-answering dataset for yes/no questions.
\item \textbf{Hellaswag} \citep{hellaswag} is a commonsense NLI dataset.
\end{itemize}

and seven knowledge-nonintensive tasks:
\begin{itemize}[leftmargin=1.3em,topsep=1pt,noitemsep]
\item \textbf{ANLI} \citep{anli} is a large-scale NLI adversarial benchmark dataset.
\item \textbf{LAMBADA} \citep{lambada} is a cloze test (word prediction) dataset.
\item \textbf{PIQA} \citep{piqa} is a physical reasoning and a corresponding benchmark dataset.
\item \textbf{RACE} \citep{race} is a large-scale reading comprehension dataset.
\item \textbf{WiC} \citep{wic} is a multilingual Word-in-Context Dataset for the evaluation of context-sensitive word embeddings.
\item \textbf{WinoGrande} \citep{winogrande} is for pronoun resolution problems.
\item \textbf{HANS} \citep{hans}  is an NLI evaluation set that tests specific hypotheses about invalid heuristics that NLI models are likely to learn.
\end{itemize}

\subsection{Evaluation Protocol} 
To evaluate autoregressive LMs on classification problems, LM Evaluation Harness Benchmark queries the LMs by concatenating the question and different candidate answers as input, comparing the probabilities of different answers, and selecting the most probable answer as LM prediction. When applying the evaluation protocol to \method, we follow the principles in \S\ref{sec:method} to separate question and answer into different chunks to avoid breaking causality.

Our \method uses the default pretraining database as the retriever.

\subsection{Fine-tuning Performance.}

Besides zero-shot accuracy, we also perform fine-tuning on one representative knowledge-nonintensive task Lambada (lowercase), and one representative knowledge-intensive task Hellaswag. 

Throughout our experiments, we fine-tune both GPT and \method for three epochs.
We use a batch size equal to 512 with a sequence length of 2048. We use the Adam optimizer (epsilon=1e-5, beta-1=0.9, beta-2=0.95) with initial lr$=$1e-5 for 530B LM, while we use lr$=$2e-5 for all other LMs. We set weight decay to 0.1 for all LMs. Our experiments are conducted on the DGX A100 servers with 8x A100 GPUs.

The fine-tuning results are shown in Table \ref{tab:finetuning}. 
We note that since Lambada (lowercase) is a more challenging dataset that consists of only lowercase samples that may hurt the retrieval quality, we observe lower accuracy of \method than GPT in the zero-shot learning setting. 
However, after fine-tuning, we observe that \method achieves better accuracy than GPT with a significant improvement margin. 
Similar observations can be found in the Hellaswag task, where \method consistently demonstrates better performance across different model sizes (Small, Medium, and XL).
This suggests that \method is better at domain-adaption after fine-tuning.

\begin{table*}[tbh!]\small
    \centering
    \resizebox{\linewidth}{!}
    {
    \begin{tabular}{lc|ll|ll|ll|ll}
    \toprule
\multicolumn{1}{l}{\multirow{2}{*}{\textbf{Tasks}}} & & \multicolumn{2}{c|}{\textbf{Small}}  & \multicolumn{2}{c|}{\textbf{Medium}}  & \multicolumn{2}{c|}{\textbf{XL}}  & \multicolumn{2}{c}{\textbf{XXL}}    \\
& & GPT & \method & GPT & \method   & GPT & \method  & GPT & \method    \\ 
\midrule
\multirow{2}{*}{\shortstack{Lambada \\(\textit{lowercase})}}  & Zero-shot     & $29.8$          & $27.0$           & $43.1$           & $43.0$           & $55.4$           & $52.5$            & $66.2$           & $65.3$          \\
                                                              & Fine-tuning   & $35.8$ \ua{6.0} & $37.2$  \ua{10.2} & $48.6$ \ua{5.5} & $50.0$  \ua{7.0} & $59.2$  \ua{3.8} & $60.0$  \ua{7.5}  & $66.8$  \ua{0.6} & $68.0$ \ua{2.7}   \\ \midrule
\multirow{2}{*}{\text{HellaSwag}}                             & Zero-shot     & $31.3$          & $36.2$           & $43.2$           & $46.2$           & $56.7$           & $59.0$            & $72.3$           & $70.6$           \\
                                                              & Fine-tuning   & $35.4$ \ua{4.1} & $40.8$  \ua{4.6} & $52.7$  \ua{9.5} & $55.1$  \ua{8.9} & $67.7$  \ua{11.0}& $68.5$  \ua{9.5}  & $75.3$  \ua{3.0} & $74.5$ \ua{3.9}  \\
\bottomrule                 
\end{tabular}
}
% \vspace{-2mm}
\caption{\small Accuracy (Acc.) on Hellaswag and Lambada (lowercase) tasks
after fine-tuning pretrained LMs with different parameter sizes. 
}
\label{tab:finetuning}
% \vspace{-3mm}
\end{table*}

\subsection{Put Retrieval Evidence in Context for GPT in zero-shot evaluation}

We have seen that retrieval significantly improves \method across different downstream tasks in the zero-shot setting. 
In this ablation study, we append the retrieval evidence of \method to the beginning of the context to see whether it can also be helpful for GPT in the zero-shot scenario. 

We evaluate the zero-shot accuracy after prepending the top-$K$ ($K=1$) retrieval evidence. The results are shown in Table \ref{tab:zeroshot-hellaswag}. We observe that directly prepending the evidence from the retrieval database messes up the GPT context in the zero-shot setting, yielding low accuracy of around $24.5\%$. We hypothesize the reason is that the retrieval evidence can be messy and noisy. Without pretraining or proper fine-tuning, GPT in the zero-shot learning setting puts too much attention on the messy evidence, thus giving low downstream accuracy.

\begin{table*}[tbh!]\small
    \centering
    % \resizebox{0.72\linewidth}{!}
    {
    \begin{tabular}{l|lc|cc|cc|cc}
    \toprule
\multicolumn{1}{l|}{\multirow{2}{*}{\textbf{Tasks}}} & \multicolumn{2}{c|}{\textbf{Small}}  & \multicolumn{2}{c|}{\textbf{Medium}}  & \multicolumn{2}{c|}{\textbf{XL}}  & \multicolumn{2}{c}{\textbf{XXL}}    \\
& GPT & GPT (retrieve) & GPT & GPT (retrieve)   & GPT & GPT (retrieve) & GPT & GPT (retrieve)   \\ 
\midrule     
  Acc. ($\uparrow$) & $31.3$          & $24.5$           & $43.2$           & $25.2$           & $56.7$           & $24.2$            & $72.3$           & $24.1$     \\
\bottomrule
\end{tabular}
}
% \vspace{-2mm}
\caption{\small  Accuracy (Acc.) on Hellaswag evaluated in the zero-shot setting.
}
\label{tab:zeroshot-hellaswag}
% \vspace{-3mm}
\end{table*}

\clearpage
\section{Details of Open-domain QA}
\label{app:training_detail_nq}

\subsection{Experimental Setup}
NQ contains questions from Google search queries and TriviaQA contains a collection of questions from trivia and quiz-league websites. Following \citet{borgeaud2022improving}, we use the processed data provided by~\citet{izacard2021leveraging} for both NQ and TriviaQA, in which each question-answer pair is accompanied by a 100-words Wikipedia passage retrieved by DPR~\citep{karpukhin2020dense}. 
We generate the answer using greedy decoding. Following the standard evaluation procedures in previous work~\cite{izacard2021leveraging, borgeaud2022improving}, Exact Match (EM) is used as our answer accuracy evaluation metric.

\subsection{Training Details}
We finetune all model parameters with the learning rate of 1e-5 for a Medium model, 3e-6 for an XL model, and 1e-6 for an XXL model.  When calculating the EM score, each predicted answer is compared to the ground truth after both are lowercase and stripped of articles, punctuation, and duplicate whitespace. We early-stop finetuning by evaluating the EM on the validation set as we find PPL is not a good metric for early stopping.

\subsection{Qualitative Study on NQ}
\label{sec: qs_nq}
Given a question, DPR retrieves a set of evidence. As $\text{RAG}_\textit{GPT}$ can only access to the first evidence, it gives the wrong answer, while \methodplus can answer correctly based on the additional evidence where the answers can be found.

\begin{table*}[ht]
    \centering
    \begin{tabular}{p{0.16\linewidth} | p{0.8\linewidth}}
      \toprule
      Question  & who invented the first home video security system \\ \midrule
      $\text{RAG}_\textit{GPT}$ & sanders associates \\
      \methodplus & marie van brittan brown \\
      Ground Truth & ['her husband Albert Brown', 'Marie Van Brittan Brown'] \\ \midrule
      Evidence 1 & 'title': 'Sanders Associates', 'text': 'Sanders Associates Sanders Associates was a defense contractor in Nashua, New Hampshire, United States, from 1951 until it was sold in 1986. It is now part of BAE Systems Electronics \& Integrated Solutions, a subsidiary of BAE Systems. It concentrated on developing and manufacturing electronic systems, notably aircraft self-protection systems, and tactical surveillance and intelligence systems. Other business areas included microwave, missile and space electronics; infrared imaging; and automated mission planning systems, with both military and commercial applications. The first home video game console was developed as a side project by engineer Ralph H. Baer and several assistants at Sanders.' \\ \midrule
      
      Evidence 2 & 'title': 'Security alarm', 'text': 'of an alarm. In some municipalities around the United States, this type of alarm verification allows the property it is protecting to be placed on a "verified response" list, allowing for quicker and safer police responses. The first video home security system was patented on December 2, 1969 to inventor Marie Brown. The system used television surveillance. To be useful, an intrusion alarm system is deactivated or reconfigured when authorized personnel are present. Authorization may be indicated in any number of ways, often with keys or codes used at the control panel or a remote panel near an entry. High-security' \\ \midrule
      
      Evidence 3 & 'title': 'Marie Van Brittan Brown', 'text': 'Marie Van Brittan Brown Marie Van Brittan Brown (October 30, 1922 – February 2, 1999) was an African-American inventor, becoming the originator of the home security system ( "") in 1966, along with her husband Albert Brown, a patent was granted in 1969. Brown was born in Jamaica, Queens, New York; she died there at the age of 76. Marie Van Brittan Brown\textbackslash{}'s father was born in Massachusetts and her mother\textbackslash{}'s roots originated from Pennsylvania. Brown and her husband lived at 151-158 \& 135th Avenue in Jamaica, Queens, New York. She worked as a nurse and her husband was an' \\ \bottomrule
    \end{tabular}
    \caption{The answer is found in Evidence 3 and \method is able to answer it correctly while $\text{RAG}_\textit{GPT}$ can only generate the answer from Evidence 1.}
    \label{tab:qualitative_study}
\end{table*}

\begin{table*}[ht]
    \centering
    \begin{tabular}{p{0.16\linewidth} | p{0.8\linewidth}}
      \toprule
      Question  & where is the hotel used in the movie the shining \\ \midrule
      $\text{RAG}_\textit{GPT}$ & estes park colorado \\
      \methodplus & stanley hotel \\
      Ground Truth & ['The Stanley Hotel'] \\ \midrule
      Evidence 1 & 'title': 'The Shining (film)', 'text': 'has become a staple of pop culture. In 2018, the film was selected for preservation in the United States National Film Registry by the Library of Congress as being "culturally, historically, or aesthetically significant." Jack Torrance arrives at the mountain-isolated Overlook Hotel, far from town, to be interviewed for the position of winter caretaker. Once hired, former teacher Jack plans to use the hotel\'s solitude to write. The hotel, built on the site of a Native American burial ground, closes during the snowed-in months. Manager Stuart Ullman tells Jack about the hotel\'s history since its 1907 construction, but he also' \\ \midrule
      
      Evidence 2 & 'title': 'The Shining (film)', 'text': 'Jan Harlan. Saint Mary Lake and Wild Goose Island in Glacier National Park, Montana was the filming location for the aerial shots of the opening scenes, with the Volkswagen Beetle driving along Going-to-the-Sun Road. The Timberline Lodge on Mount Hood in Oregon was filmed for a few of the establishing shots of the fictional Overlook Hotel; notably absent in these shots is the hedge maze, something the Timberline Lodge does not have. Outtakes of the opening panorama shots were later used by Ridley Scott for the closing moments of the original cut of the film "Blade Runner" (1982). "The Shining"' \\ \midrule
      
      Evidence 3 & 'title': 'The Shining (film)', 'text': 'order, he used several stages at EMI Elstree Studios in order to make all sets available during the complete duration of production. The set for the Overlook Hotel was at the time the largest ever built at Elstree, including a life-size re-creation of the exterior of the hotel. In February 1979, the set at Elstree was badly damaged in a fire, causing a delay in the production. While most of the interior shots, and even some of the Overlook exterior shots, were shot on studio sets, a few exterior shots were shot on location by a second-unit crew headed by' \\ \midrule
        
      Evidence 4 &  'title': 'The Shining (film)', 'text': 'end of the film and Jack\'s repeated claims to have "not just a deja vu". The film is even more focused on Jack (as opposed to Danny) than the novel. The room number 217 has been changed to 237. Timberline Lodge, located on Mt. Hood in Oregon, was used for the exterior shots of the fictional Overlook Hotel. The Lodge requested that Kubrick not depict Room 217 (featured in the book) in "The Shining", because future guests at the Lodge might be afraid to stay there, and a nonexistent room, 237, was substituted in the film. Contrary to the hotel\'s' \\ \midrule

      Evidence 5 &  'title': 'The Stanley Hotel', 'text': 'main building which adorned the lawn of the Overlook Hotel in the series can be viewed in the basement of the Stanley. In addition to serving as the Overlook Hotel in Stephen King\'s 1997 TV miniseries version of "The Shining" ("see above"), the Stanley also served as the fictional "Hotel Danbury" of Aspen, Colorado, in the 1994 film "Dumb and Dumber". From 2013 to 2015, the hotel property hosted the Stanley Film Festival, an independent horror film festival operated by the Denver Film Society, held in early May. The festival featured screenings, panels, student competitions, audience awards and receptions. The' \\ \bottomrule
    \end{tabular}
    \caption{The answer is found in Evidence 5 and \method is able to answer it correctly while $\text{RAG}_\textit{GPT}$ cannot.}
    \label{tab:qualitative_study}
\end{table*}

\end{document}